%% file: main.tex
\documentclass[10pt,twocolumn,letterpaper]{article}

\usepackage[pagenumbers]{cvpr} 

\input{preamble}
\definecolor{cvprblue}{rgb}{0.21,0.49,0.74}
\usepackage[pagebackref,breaklinks,colorlinks,allcolors=cvprblue]{hyperref}

\def\confName{CVPR}
\def\confYear{2027}

\title{\method for Efficient MLLMs}

\author{Bryan A. Plummer
\and Shan Lu
\and Keval Doshi
\and Noranart Vesdapunt
\and Wei Wang
\and Jim Thomas 
\and Kah Kuen Fu \\
\hspace{-4cm}Amazon
}

\usepackage{algorithm}
\usepackage{algpseudocode}
\usepackage{multirow}

\newcommand{\method}{Balanced Twin Weight Factorization\xspace}
\newcommand{\methodabbr}{BTWF\xspace}
\newcommand{\task}{Weight Factorization\xspace}
\newcommand{\taskabbr}{WF\xspace}
\newcommand{\flops}{FLOPs\xspace}
\newcommand{\balanced}{Factor Component Balancing\xspace}
\newcommand{\balancedabbr}{FCB\xspace}
\newcommand{\downstream}{Block Contribution Balancing\xspace}
\newcommand{\downstreamabbr}{BCB\xspace}
\newcommand{\overparam}{Overparameterized Groups\xspace}
\newcommand{\overparamabbr}{OG\xspace}

\begin{document}
\maketitle
\input{sec/0_abstract}    
\input{sec/1_intro}
\input{sec/2_related_work}

\input{sec/3_method}

\input{sec/4_experiments}

\input{sec/5_conclusion}
{
    \small
    \bibliographystyle{ieeenat_fullname}
    \bibliography{main}
}

\input{sec/X_suppl}

\end{document}

%% file: sec/0_abstract.tex
\begin{abstract}
\task (\taskabbr) methods aim to reduce a model's latency, memory, or both.  These typically decompose a pretrained model into two or more factors that approximate the original model weights with fewer parameters and floating point operations (\flops).   However, translating fewer \flops into reduced latency is challenging for \taskabbr methods in practice due to overhead introduced by replacing a single large weight matrix with multiple smaller ones.  To address this, we propose \method (\methodabbr), which uses the unique structure of factorized models to simultaneously lower latency and increase learning capacity.  Specifically, \methodabbr operates in the shared basis \taskabbr setting, where two or more layers are decomposed into a shared basis matrix and a layer-specific mixing matrix.  If these shared layers have the same input (\ie, are twins), then we need only perform this shared forward pass once, reducing inference time.  However, twin layers are often of varying sizes, and methods from prior work tend to favor the larger layer.  In contrast, our approach boosts performance by combining \balanced (\balancedabbr), which ensures every shared member is well represented in the selected factors, with \downstream (\downstreamabbr), which considers downstream model effects.   Further, we show that an overparameterized \methodabbr, where some layer's factors result in higher FLOPs than the original during training, can boost task performance without additional resources at test time.  Experiments on six models over nine benchmarks show \methodabbr obtains a 2-6\% gain over the state-of-the-art without expensive finetuning (a 1-3.5\% gain with it) at comparable computational costs.
\end{abstract}

%% file: sec/1_intro.tex
\begin{figure}[t]
  \centering
  \includegraphics[width=0.92\linewidth]{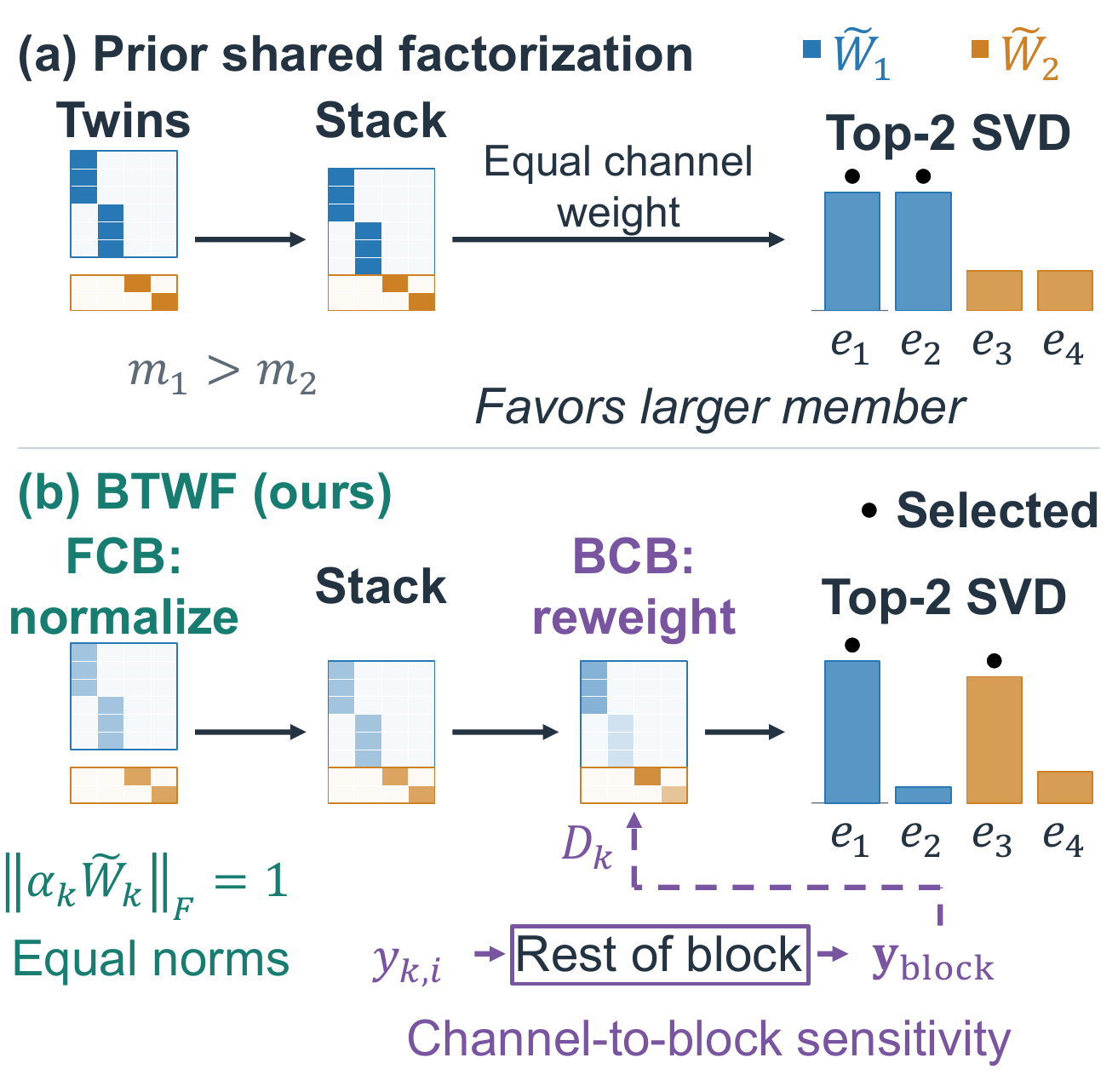}
  \vspace{-4mm}
  \caption{\textbf{Illustrative example selecting a shared rank-two subspace.} Blue and orange denote whitened twin members $\tilde W_k=W_kL$, whereas black dots mark selected directions. \textbf{(a)} Prior work directly decomposes twin member's concatenation~\cite{wang2025basissharing,an2026glowq,lu2026skipcat}, resulting in equal weighting of each channel.  Thus, the larger member with more channels has a higher likelihood to have their factors selected (selecting $\{e_1,e_2\}$ in this example). \textbf{(b)} our \methodabbr approach addresses the large-member bias via per-member norm balancing with \balancedabbr.  Then \downstreamabbr improves dense model fidelity by reweighting channels  according to their sensitivity to the containing block's output (resulting in $\{e_1,e_3\}$ being chosen).}
  \label{fig:motivation}
  \vspace{-2mm}
\end{figure}

\begin{figure*}[t]
  \centering
  \includegraphics[width=0.66\textwidth]{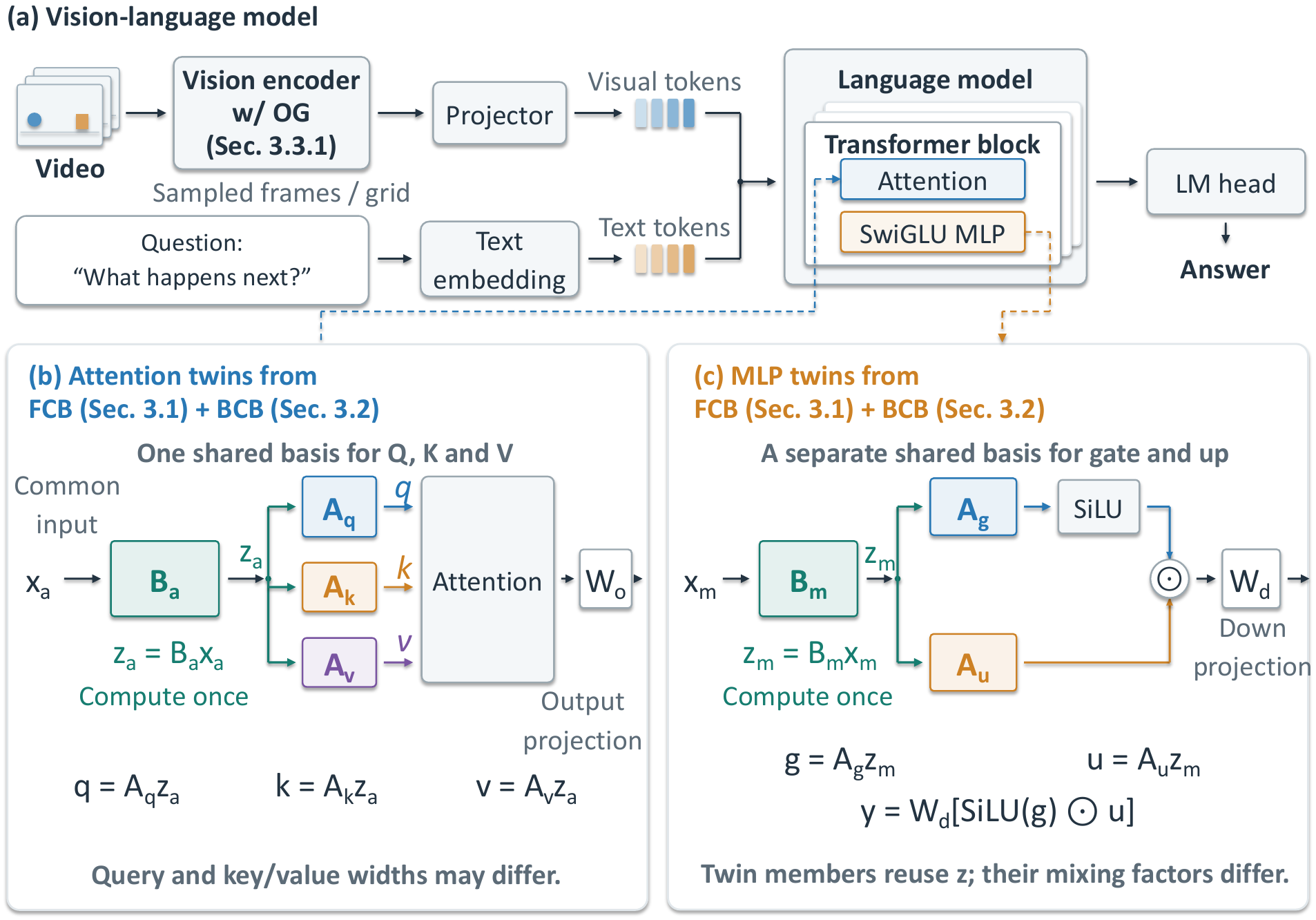}
  \vspace{-2mm}
  \caption{\textbf{Example implementation of our \methodabbr shared  factorization on an InternVL3-2B~\cite{zhu2025internvl3} backbone.} \textbf{(a)} Shows the overall structure of the vision-language model, which remains unchanged. Instead, we factorize layers using \balanced (\balancedabbr) described in \cref{sec:balanced} and \downstream (\downstreamabbr) discussed in \cref{sec:downstream}.  These can either use \overparam (\overparamabbr) from \cref{sec:opgroups} that use large ranks to add model capacity (applied to the vision encoder in this example) or use a small rank to reduce memory and latency on twin groups within a transformer block. \textbf{(b)} shows attention twin groups where query, key, and value projections reuse $z_{\mathrm{a}}=B_{\mathrm{a}}x_{\mathrm{a}}$, computed once, with separate mixing factors $A_{\mathrm{q}}$, $A_{\mathrm{k}}$, and $A_{\mathrm{v}}$. The attention operation is general and permits different query and key/value widths. 
  \textbf{(c)} illustrates that gate and up projections in a SwiGLU MLP similarly can reuse $z_{\mathrm{m}}=B_{\mathrm{m}}x_{\mathrm{m}}$. }
  \vspace{-2mm}
  \label{fig:method}
\end{figure*}

\section{Introduction}

\task (\taskabbr) methods reduce a model's latency and memory by decomposing a large weight matrix into two or more smaller ones~\cite{plummerNPAS2022,phamMixtureGrowth2024,teterwakSuperWeight2024,tasnimRECAST2025,TasnimCRISP2026,savarese2018learning,Bagherinezhad_2017_CVPR,wang2025svdllm,wang2025basissharing,Li_2026_CVPR_Adasvd}. 
However, the latency drop realized by these methods is less than their reduction in the number of floating-point operations (\flops) due to the overhead required to launch a sequence of small matrix multiplications.  In fact, many optimization methods explicitly avoid this additional cost by reducing the number of matrix multiplications in a forward pass (\eg, by fusing key-query-value attention)~\cite{pope2023efficiently,aminabadi2022deepspeedinference}.  Prior work addresses this by creating a shared basis factor between twin layers that see the same input and, thus, a shared basis need only be computed once~\cite{an2026glowq,lu2026skipcat}.  Then, unshared mixing factors capture layer-specific information. However, as illustrated in \cref{fig:motivation}(a), when shared layers are of unequal sizes (\eg, GQA~\cite{ainslie2023gqa}, MQA~\cite{shazeer2019mqa}, and MLA~\cite{deepseekai2024v2} attention as well as gate and up projections of a SwiGLU MLP~\cite{shazeer2020glu}), their simple concatenation then decomposition results in favoring larger layers.

This paper directly addresses these practical limitations via our \method (\methodabbr) approach.  Specifically, as shown in \cref{fig:motivation}(b), when selecting what factors to keep we introduce two benefits over prior \taskabbr work~\cite{wang2025basissharing,an2026glowq,lu2026skipcat}.  First, \balanced (\balancedabbr) ensures each twin member is well represented by scaling shared basis members by their Frobenius norms before decomposition.  This helps address a bias in prior work towards selecting factors that benefit larger members more than smaller ones.  Second, we propose \downstream (\downstreamabbr), which weights factors by their contribution to block-level output error. This helps us retain factors that may not contribute as much to reconstructing the original model's weights, but can help keep information more useful for future layers.  

Despite the more effective factors provided by \methodabbr, our experiments still show that some layers rapidly degrade performance as their rank shrinks. This echoes similar observations in related work that highlighted unequal effects from removing similar weight counts for different layers~\cite{plummerNPAS2022,teterwakSuperWeight2024,liang2025beyond,xiong2026d2prune,mugnaini2025amp,coello2026putri,yang2025prunellm,kwon2022fast}. To address this, we show that  \methodabbr can also be used to create \overparam (\overparamabbr) on these low-rank sensitive layers that use more total parameters than the layers they are sourced from, effectively increasing model capacity instead of reducing it.  At inference time these \overparamabbr factors are baked back to the original layer size (\ie, they require no more resources than the original model after training).  Thus, this shows our single framework can simultaneously be used to reduce latency and boost task performance.  See \cref{fig:method} for an example of how \methodabbr is used. Experiments on six models of various sizes and families evaluated on nine diverse benchmarks on image-language and video-language tasks show that \methodabbr outperforms the state-of-the-art when controlling for \flops, latency, and memory.

In summary, our contributions are:
\begin{itemize}[nosep,leftmargin=*]
    \item We propose \methodabbr, a shared \taskabbr approach that uses \balancedabbr to minimize the bias favoring larger layers suffered by prior work, resulting in gains of 2-6\% without expensive finetuning and 1-3.5\% with it.
    \item We show that we can more effectively reconstruct layer weights thanks to \downstreamabbr, which helps retain more informative factors by considering the effect on its block output.
    \item We demonstrate \methodabbr's ability to serve as a drop-in replacement for methods that compose with weight pruning, token pruning, and quantization for further gains.
\end{itemize}

%% file: sec/2_related_work.tex
\section{Related Work}

Shared \taskabbr methods are closest to ours, where prior work did not account for the varying layer sizes in their reconstruction~\cite{an2026glowq,lu2026skipcat,wang2025basissharing}, forwent the inference benefits of twin layers discussed earlier by sharing only a portion of the basis factor~\cite{ji2025mha2mla,meng2025transmla}, or trained from scratch instead of adapting pretrained models~\cite{plummerNPAS2022,teterwakSuperWeight2024,yu2018slimmable,Yu_2019_ICCV,deepseekai2024v2,guo2020expandnets}.   There are also some \taskabbr methods that focused on other tasks like model compression~\cite{wang2025svdllm,wang2025basissharing,plummerNPAS2022,teterwakSuperWeight2024,yuan2023asvd,wang2025dobisvd,Li_2026_CVPR_Adasvd}, continual learning~\cite{tasnimRECAST2025,TasnimCRISP2026}, or reducing training time~\cite{phamMixtureGrowth2024}. However, these methods did not reduce latency and/or considered how retained factors contribute on its block output.

Related tasks that reduce a model's memory and latency include weight pruning~\cite{lecun1989obd,han2015learning,li2017pruning,molchanov2017pruning,theis2018fisher,kwon2022fast,frantar2023sparsegpt,ma2023llmpruner,sun2024wanda,an2024flap,sung2024ecoflap,men2025shortgpt,mugnaini2025amp,yang2025prunellm,guo2025slimllm,qu2025geta,liang2025beyond,xiong2026d2prune,zhang2026salr,wu2026comp,madinei2026interlace,tan2026prism,yamamoto2026mop,coello2026putri,fang2023depgraph,li2025trthepruner,Ketanbhai2026topk,MinSeok2026PORTA,michel2019sixteen,voita2019analyzing,yin2024owl,farina2024multiflow,ashkboos2024slicegpt,shi2023upop}, attention compression~\cite{michel2019sixteen,voita2019analyzing,liang2025beyond,ma2023llmpruner,an2024flap,guo2025slimllm,mugnaini2025amp,coello2026putri,yang2025prunellm}, model quantization~\cite{lin2024awq,frantar2023gptq,yao2022zeroquant,dettmers2022llmint8,jacob2018quantization,rastegari2016xnornet,han2016deepcompression,xiao2023smoothquant,chiang2026uniql}, or knowledge distillation~\cite{gu2024minillm,touvron2021deit,jiao2020tinybert,sanh2019distilbert,zagoruyko2017attention,hinton2015distilling}.  However, prior work has shown \taskabbr methods can be combined with these tasks for additional gains~\cite{an2026glowq,lu2026skipcat,TasnimCRISP2026,plummerNPAS2022,chang2025palu}. We will show \methodabbr can be used as a drop-in replacement to boost performance further.  

Token pruning~\cite{wang2021spatten,yin2022avit,liang2022evit,rao2021dynamicvit,kim2022ltp,bolya2023tome,chen2024fastv} and mixture-of-experts~\cite{dai2024deepseekmoe,jiang2024mixtral,zoph2022stmoe,riquelme2021vmoe,zhou2022expertchoice,du2022glam,fedus2022switch,lepikhin2021gshard,shazeer2017outrageously} are examples of tasks that focus on reducing inference time rather than memory (ours reduces both). Prior work has shown they can also be combined with methods like \methodabbr for an additional boost~\cite{wu2026comp}. As with the weight pruning and quantization methods, our experiments report gains from using \methodabbr as drop-in replacement in this setting too.

%% file: sec/3_method.tex
\section{\method (\methodabbr)}

Given a set of pretrained weights
$\mathcal{W}=\{W_k\in\mathbb{R}^{m_k\times n}\}_{k=1}^{K}$ whose members share the input width
$n$ but may differ in output width $m_k$ (\eg, under GQA attention~\cite{ainslie2023gqa}, $m_{\mathrm{q}}>m_{\mathrm{k}}=m_{\mathrm{v}}$), our goal is a
rank-$r$ factorization that approximates every member of $\mathcal{W}$ with a shared input
factor, \ie,
\begin{align}
  W_k \;\approx\; A_k B, \text{where}\hspace{2.5cm} \nonumber\\ 
  A_k\in\mathbb{R}^{m_k\times r}\ \text{(per member)},\quad
  B\in\mathbb{R}^{r\times n}\ \text{(shared)}.
  \label{eq:shared}
\end{align}
$B$ is a shared basis for the group's input, whereas (unshared) mixing matrix
$A_k$ captures unique member information. 

Following~\cite{lu2026skipcat}, we group
layers that read the same input tensor and factorize other layers as singletons. \Eg, query, key and value projections all read the normalized hidden state entering
attention~\cite{vaswani2023attentionneed}, as do the gate and up projections of a SwiGLU
MLP~\cite{shazeer2020glu}. Thus, the projection $z=Bx$ is computed once and reused by
every member, $y_k=A_kz$. \methodabbr makes two primary improvements over prior
work~\cite{lu2026skipcat,wang2025basissharing,an2026glowq,wang2025svdllm,Li_2026_CVPR_Adasvd}. First,
\cref{sec:balanced} discusses \balancedabbr, which balances shared factors between members of
different sizes. Second, \cref{sec:downstream} describes \downstreamabbr, which considers a factor's affect its block's output. 

\subsection{\balanced (\balancedabbr)}
\label{sec:balanced}
In this section we describe \balancedabbr, which balances the contribution of each member to the
reconstruction loss; it is combined with \downstreamabbr in \cref{sec:downstream}. \taskabbr methods
like ours create the factors of \cref{eq:shared} from a pretrained model by minimizing
reconstruction error, \eg,
\begin{equation}
  \label{eq:reconstruct}
  \min_{V\in\mathbb{R}^{n\times r},\,\{A_k\}}\;
  \sum_{k=1}^{K}\big\|W_k - A_k V^\top\big\|_F^2 ,
\end{equation}
where $\hat W_k=A_kV^\top$ and, without further weighting, the shared factor is simply
$B=V^\top$. Some methods optimize \cref{eq:reconstruct} directly with gradient
descent~\cite{tasnimRECAST2025,TasnimCRISP2026}, but this often results in significant model drift.

Prior work addresses this by weighting the error with calibration activations
$X\in\mathbb{R}^{n\times N}$ entering $\mathcal{W}$~\cite{wang2025svdllm,Li_2026_CVPR_Adasvd}. Let
$S=XX^\top+\lambda I$ be the ridge-regularized input second moment and $L$ its Cholesky factor,
$S=LL^\top$. Then $\|(W_k-\hat W_k)X\|_F^2\approx\|(W_k-\hat W_k)L\|_F^2$ (exact for
$\lambda=0$), so this is a plain Frobenius objective on
\emph{whitened} members $\tilde W_k=W_kL$. Under an orthonormality constraint $V^\top V=I$ its optimum $V$ is the top-$r$ eigenspace of
$\sum_k\tilde W_k^\top\tilde W_k$. \Ie, the top-$r$ right singular subspace of the vertically
stacked $[\tilde W_1;\dots;\tilde W_K]$ with coefficients $A_k=\tilde W_kV$. Undoing the
whitening gives \cref{eq:shared}'s factors,
\begin{equation}
  B \;=\; V^\top L^{-1}, \qquad A_k \;=\; W_k L V ,
\end{equation}
so $\hat W_k=A_kB=W_kLVV^\top L^{-1}$, and $(W_k-\hat W_k)L=\tilde W_k-A_kV^\top$.

This is the general approach
of~\cite{lu2026skipcat,wang2025basissharing,an2026glowq} that directly extends the unshared
factorization of \citet{wang2025svdllm} (where $\mathcal{W}$ is a singleton) to a shared factor.
However, the plain sum in \cref{eq:reconstruct} has  each member pulling on
the shared subspace in proportion to its energy $\|\tilde W_k\|_F^2$, which grows with
its output width $m_k$ and with the scale of its weights and outputs. Thus, large members dominate $V$, and small members (\eg, the key and value projections under GQA) are reconstructed
with much larger relative error.

\balancedabbr helps members balance their contribution regardless of $m_k$ or scale by normalizing each whitened target by its Frobenius norm. With
$\alpha_k=\|\tilde W_k\|_F^{-\gamma}$, our objective is
\begin{equation}
  \label{eq:balanced}
  \min_{V\in\mathbb{R}^{n\times r},\,V^\top V=I,\,\{A_k\}}\;
  \sum_{k=1}^{K}\alpha_k^{2}\,\big\|\tilde W_k-A_kV^\top\big\|_F^2 ,
\end{equation}
whose optimum $V$ is the top-$r$ right singular subspace of
$[\alpha_1\tilde W_1;\dots;\alpha_K\tilde W_K]$. The coefficients remain $A_k=\tilde W_kV$, so
the weights act only on the choice of the shared subspace. When $\gamma=1$ each member
enters through its relative whitened error
$\|(W_k-\hat W_k)L\|_F^2/\|\tilde W_k\|_F^2$, giving equal weight between members.

This approach is
similar to Balanced Key--Value (BKV) scaling that helped
convert a pretrained model from GQA to MLA attention \cite{meng2025transmla}. BKV rescales key activations by the
ratio of the mean per-token $\ell_2$ norms of the keys and values.  Thus, one member matches the
other and the scaling is defined \textit{only for a pair}. In contrast, \balancedabbr uses a whitened
Frobenius norm $\|\tilde W_k\|_F\approx(\sum_t\|W_kx_t\|_2^2)^{1/2}$, our objective's metric, and normalizes any number of members to a common scale to support larger groups.  Further, when $\gamma \neq 1$, \balancedabbr allows for deliberate bias either towards larger ($\gamma < 1$) or smaller ($\gamma > 1$) weight norms, which is set using validation data (see  \cref{sec:balanced_study}). In summary, \cref{eq:balanced} trades total reconstruction error for more balance
across members.  However, this does not consider each factor's importance, which we address in the next section.

\subsection{\downstream (\downstreamabbr)}
\label{sec:downstream}

The reconstruction loss described at the end of \cref{sec:balanced} assumes that reconstructing
the activations of each member of $\mathcal{W}$ would produce the best factorization. However,
prior work in weight pruning showed that some activations can be removed entirely with little
loss in performance, if any~\cite{lecun1989obd,MinSeok2026PORTA,sun2024wanda,an2024flap}. Our
\downstreamabbr module takes advantage of this observation by focusing reconstruction on
activations that are more likely to affect performance.

More formally, \cref{eq:balanced} weights every output row of a member equally, \ie, an error in
row $i$ of $\hat W_k$ is penalized the same whether the activation it produces feeds a large or a
negligible part of the block's output. Instead, \downstreamabbr weights each row by a first-order
estimate of how strongly an error in that activation propagates to the output of the block that
contains $\mathcal{W}$. Let $D_k=\operatorname{diag}(\delta_k)$ with
$\delta_k\in\mathbb{R}_{>0}^{m_k}$ hold these \emph{row weights} (defined below). We solve
\begin{equation}
  \label{eq:downstream}
  \min_{V\in\mathbb{R}^{n\times r},\,V^\top V=I,\,\{A_k\}}\;
    \sum_{k=1}^{K}\alpha_k^{2}\,\big\|D_k\big(\tilde W_k-A_kV^\top\big)\big\|_F^2 ,
\end{equation}
where the whitening $\tilde W_k=W_kL$ weights the error by the member's \emph{inputs} (\cref{sec:balanced}) and $D_k$ weights it by where its \emph{outputs} go. Writing $e_{k,i,t}$ for the error of row $i$ on calibration
token $t$, \cref{eq:downstream} approximates
$\sum_k\alpha_k^2\sum_{i,t}\delta_{k,i}^2\,e_{k,i,t}^2$, the first-order squared error at the
block output, with cross-row terms dropped and each row's sensitivity replaced by its
calibration average.

For a fixed $V$ the problem still separates by row, and because $D_k$ is diagonal with nonnegative entries, each row's weight rescales its residual
without moving its minimizer. \Ie, the least-squares coefficients are those of the
unweighted problem, $A_k=\tilde W_k V$.  Like $\alpha_k$, the row weights
therefore act \emph{only} on the choice of the shared subspace, which is now the top-$r$ right
singular subspace of
\begin{equation}
  \label{eq:downstream-stack}
  \big[\alpha_1D_1\tilde W_1;\ \dots;\ \alpha_KD_K\tilde W_K\big].
\end{equation}
The balancing weights are computed from the unweighted whitened targets,
$\alpha_k=\|\tilde W_k\|_F^{-\gamma}$, so \balancedabbr still sets the share of each member and
\downstreamabbr redistributes rank among the rows within it. \cref{alg:twin} summarizes our \methodabbr factorization procedure. See  \cref{sec:rowweights} for examples of how to compute $\delta$ for different operations.

Our \downstreamabbr module is similar to prior work that balanced factors by their group output rather than the transformer block~\cite{lin2025modegpt} or used a task loss~\cite{hsu2022fwsvd}.  However,  \cref{sec:downstream_study} shows these are less effective than \downstreamabbr.

\begin{algorithm}[t]
\caption{\methodabbr for one shared group $\mathcal{W}=\{W_k\}_{k=1}^{K}$ with a common input}
\label{alg:twin}
\begin{algorithmic}[1]
\Require members $W_k\in\mathbb{R}^{m_k\times n}$; calibration inputs $X\in\mathbb{R}^{n\times N}$;
         rank $r$ (see \cref{sec:setr})
\State $S \gets XX^\top+\lambda I$;\quad $L \gets \operatorname{chol}(S)$;\quad
       $\tilde W_k \gets W_k L$ for all $k$
       \Comment{whiten (\cref{sec:balanced})}
\State $\alpha_k \gets \|\tilde W_k\|_F^{-\gamma}$ for all $k$ \Comment{\balancedabbr (\cref{sec:balanced})}
\State $D_k \gets \operatorname{diag}(\delta_k)$, examples in \cref{sec:rowweights} \Comment{\downstreamabbr (\cref{sec:downstream})}
\State $G \gets \sum_{k=1}^{K}\alpha_k^2\,\tilde W_k^\top D_k^2\,\tilde W_k$
       \Comment{Gram of $[\alpha_1D_1\tilde W_1;\dots;\alpha_KD_K\tilde W_K]$}
\State $V \gets$ top-$r$ eigenvectors of $G$ \Comment{$V\in\mathbb{R}^{n\times r}$, $V^\top V=I$}
\State $A_k \gets \tilde W_k V$ for all $k$ \Comment{weights choose $V$ only}
\State $B \gets V^\top L^{-1}$ \Comment{triangular solve $L^\top B^\top=V$}
\State $\Lambda \gets$ diagonal balancing of $\|B_{i,:}\|$ against the geometric mean over $k$
       of $\|A_{k,:,i}\|$
\State $A_k \gets A_k\Lambda$ for all $k$;\quad $B \gets \Lambda^{-1}B$
       \Comment{every $A_kB$ unchanged}
\State \Return $B$ (shared; $z=Bx$ computed once per token), $\{A_k\}_{k=1}^{K}$
\end{algorithmic}
\end{algorithm}

\begin{table*}[t]
\centering
\caption{Results on criteria-only structured weight pruning/\taskabbr results averaged over three runs on InternVL3-2B~\cite{zhu2025internvl3} using only 256 samples from CLEVRER~\cite{yi2020clevrer} (one of the three Seen datasets) for any calibration.  For every method, all layers in the network must reduce the same number of \flops (\ie, uniform pruning/\taskabbr).  Our \methodabbr approach reports a 2-3\% overall gain over prior work. See \cref{tab:strategy} for results that include pruning/\taskabbr strategy (\ie, unequal budgets across layers).}
\vspace{-2mm}
\begin{tabular}{lcccccc}
\toprule
\flops Budget & \multicolumn{3}{c}{30\%} & \multicolumn{3}{c}{50\%} \\
\cmidrule(lr){2-4} \cmidrule(lr){5-7}
Method & Seen & Unseen & All & Seen & Unseen & All \\
\midrule
Wanda~\cite{sun2024wanda} & 0.8{\scriptsize$\pm$0.3} & 1.2{\scriptsize$\pm$0.5} & 1.1{\scriptsize$\pm$0.4} & 1.2{\scriptsize$\pm$0.2} & 2.0{\scriptsize$\pm$0.2} & 1.7{\scriptsize$\pm$0.2} \\
Putri~\cite{coello2026putri} & 2.4{\scriptsize$\pm$1.8} & 3.2{\scriptsize$\pm$2.7} & 3.0{\scriptsize$\pm$2.3} & 1.8{\scriptsize$\pm$0.9} & 2.7{\scriptsize$\pm$1.5} & 2.4{\scriptsize$\pm$1.3} \\
Wanda~\cite{sun2024wanda}+Putri~\cite{coello2026putri} & 25.2{\scriptsize$\pm$1.0} & 24.0{\scriptsize$\pm$0.3} & 24.4{\scriptsize$\pm$0.5} & 3.0{\scriptsize$\pm$1.9} & 3.2{\scriptsize$\pm$2.3} & 3.1{\scriptsize$\pm$2.2} \\
PruneLLM~\cite{yang2025prunellm} & 1.2{\scriptsize$\pm$0.4} & 2.3{\scriptsize$\pm$0.5} & 1.9{\scriptsize$\pm$0.5} & 0.2{\scriptsize$\pm$0.1} & 0.7{\scriptsize$\pm$0.6} & 0.5{\scriptsize$\pm$0.5} \\
D$^2$Prune~\cite{xiong2026d2prune} & 24.4{\scriptsize$\pm$1.0} & 23.2{\scriptsize$\pm$0.4} & 23.6{\scriptsize$\pm$0.6} & 17.5{\scriptsize$\pm$0.1} & 20.6{\scriptsize$\pm$0.5} & 19.6{\scriptsize$\pm$0.3} \\
RCPU~\cite{haruta2026rcpu} & 30.0{\scriptsize$\pm$0.4} & 28.5{\scriptsize$\pm$1.1} & 29.0{\scriptsize$\pm$0.7} & 2.1{\scriptsize$\pm$0.2} & 3.1{\scriptsize$\pm$0.5} & 2.8{\scriptsize$\pm$0.4} \\
SVD-LLM~\cite{wang2025svdllm} & 25.8{\scriptsize$\pm$0.6} & 23.5{\scriptsize$\pm$0.5} & 24.3{\scriptsize$\pm$0.5} & 16.3{\scriptsize$\pm$1.8} & 14.1{\scriptsize$\pm$2.8} & 14.9{\scriptsize$\pm$2.5} \\
Basis Sharing~\cite{wang2025basissharing} & 24.9{\scriptsize$\pm$0.3} & 21.9{\scriptsize$\pm$0.6} & 22.9{\scriptsize$\pm$0.4} & 14.9{\scriptsize$\pm$4.9} & 13.3{\scriptsize$\pm$6.2} & 13.8{\scriptsize$\pm$5.7} \\
Cat~\cite{lu2026skipcat} & 30.5{\scriptsize$\pm$0.8} & 26.9{\scriptsize$\pm$1.2} & 28.1{\scriptsize$\pm$1.1} & 20.8{\scriptsize$\pm$0.9} & 20.0{\scriptsize$\pm$0.2} & 20.3{\scriptsize$\pm$0.4} \\
\methodabbr w/o \downstreamabbr & 33.6{\scriptsize$\pm$0.4} & 29.6{\scriptsize$\pm$0.9} & 30.9{\scriptsize$\pm$0.6} & \textbf{23.4}{\scriptsize$\pm$0.6} & 20.7{\scriptsize$\pm$0.5} & 21.6{\scriptsize$\pm$0.4} \\
\methodabbr w/o \balancedabbr  & \textbf{34.5}{\scriptsize$\pm$0.7} & 30.7{\scriptsize$\pm$0.8} & 32.0{\scriptsize$\pm$0.7} & 22.8{\scriptsize$\pm$0.4} & 21.3{\scriptsize$\pm$0.1} & 21.8{\scriptsize$\pm$0.2} \\
\methodabbr (both, ours) & 34.4{\scriptsize$\pm$1.0} & \textbf{31.2}{\scriptsize$\pm$0.9} & \textbf{32.2}{\scriptsize$\pm$0.9} & 23.3{\scriptsize$\pm$0.7} & \textbf{21.6}{\scriptsize$\pm$0.5} & \textbf{22.2}{\scriptsize$\pm$0.5} \\
\bottomrule
\end{tabular}
\vspace{-2mm}
\label{tab:criteria}
\end{table*}

\subsection{Setting Rank $r$}
\label{sec:setr}
The rank is set to capture a target budget $\Omega$ (\eg, \flops, wall time, memory).  \Eg, given a \flops budget $\Omega_f$, the per token costs in the shared factorization of
\cref{eq:shared} for a group
$\mathcal{W}=\{W_k\in\mathbb{R}^{m_k\times n}\}_{k=1}^{K}$ (a singleton when $K=1$) with
$M=\sum_k m_k$ costs $2r(n+M)$ against $2nM$ dense. Every
group receives the same fraction of its break-even rank,
\begin{equation}
  \label{eq:rank}
  r=\big\lfloor \rho\,\frac{nM}{(n+M)}\big\rfloor ,
\end{equation}
where ratio $\rho$ is solved by bisection on the measured whole-model \flops, so
every compressed model meets $\Omega_f$. A group is charged $\min\!\big(2r(n+M),\,2nM\big)$,
since factorizations above break-even can be folded into dense weights.

\subsubsection{\overparam (\overparamabbr)}
\label{sec:opgroups}
A chosen set $\mathcal{P}$ of groups can instead be held at dense cost with \emph{more}
capacity than their dense layers,
\begin{equation}
  \label{eq:overparam}
  r_{\mathcal{W}}=\big\lfloor \mu\,\frac{nM}{(n+M)}\big\rfloor,\qquad \mu\ge 1,\ \mathcal{W}\in\mathcal{P},
\end{equation}
so factors hold $\mu$ times the dense parameter count. Each $A_kB$ is multiplied out once
after recovery, so these groups add no inference \flops, and $\rho$ is re-solved over the
remaining layers of both towers to meet the same budget as $\mu=1$. A budget is
feasible only if those linears can absorb it, \eg, holding every MLP of InternVL3-2B~\cite{zhu2025internvl3} at dense cost
leaves only enough for the 10\% rate.

The shared basis spans at most $\min(n,M)$ directions, so the products $A_kB$ are
unconstrained once $r_{\mathcal{W}}\ge\min(n,M)$, \ie, for
\begin{equation}
  \label{eq:overparam-exact}
  \mu\;\ge\;\mu^\ast=\frac{\min(n,M)\,(n+M)}{nM},
\end{equation}
where the initialization of \cref{sec:balanced,sec:downstream} reproduces $W_k$ exactly.
For InternVL3-2B~\cite{zhu2025internvl3}, $\mu^\ast$ is $1.75$ for the language model's query/key/value group
($n=1536$, $M=2048$), $1.09$ for its gate/up group, $1.17$ for its down projection, $1.25$ for
the vision MLP, and $1.33$ for the vision query/key/value projection. Below $\mu^\ast$ the
shared row space constrains members, while at or above it the group starts as dense layers in a factored parameterization. Undetermined rank directions are
initialized to train rather than left inert.

%% file: sec/4_experiments.tex
\section{Experiments}

\begin{table*}[t]
\centering
\small
\setlength{\tabcolsep}{3pt}
\caption{Results on structured weight pruning/\taskabbr results on an InternVL3-2B~\cite{zhu2025internvl3} backbone averaged over three runs using only 256 samples from CLEVRER~\cite{yi2020clevrer} (one of the three Seen datasets) for any calibration.  Each method uses the budget allocation strategy from their paper to achieve the same \flops budget as used in \cref{tab:criteria}. Our approach includes \overparam (\overparamabbr) described in \cref{sec:opgroups} for some layers, with any remaining layers using a higher rate of uniform \taskabbr to obtain the total \flops budget.  Our \methodabbr approach reports a 3-6\% overall gain over prior work.}
\vspace{-2mm}
\begin{tabular}{llccccccccc}
\toprule
\multicolumn{2}{l}{\flops Budget} & \multicolumn{3}{c}{10\%} & \multicolumn{3}{c}{30\%} & \multicolumn{3}{c}{50\%} \\
\cmidrule(lr){3-5} \cmidrule(lr){6-8} \cmidrule(lr){9-11}
Weight Selector & Budget Allocator & Seen & Unseen & All & Seen & Unseen & All & Seen & Unseen & All \\
\midrule
LLM-Pruner~\cite{ma2023llmpruner} & LLM-Pruner~\cite{ma2023llmpruner} & 35.4{\scriptsize$\pm$0.2} & 41.8{\scriptsize$\pm$0.1} & 39.6{\scriptsize$\pm$0.1} & 3.7{\scriptsize$\pm$0.6} & 5.1{\scriptsize$\pm$0.3} & 4.6{\scriptsize$\pm$0.3} & 1.3{\scriptsize$\pm$0.4} & 0.6{\scriptsize$\pm$0.1} & 0.8{\scriptsize$\pm$0.2} \\
SlimLLM~\cite{guo2025slimllm} & SlimLLM~\cite{guo2025slimllm} & 46.0{\scriptsize$\pm$0.2} & 51.9{\scriptsize$\pm$0.3} & 49.9{\scriptsize$\pm$0.2} & 8.5{\scriptsize$\pm$3.4} & 10.4{\scriptsize$\pm$2.7} & 9.8{\scriptsize$\pm$2.9} & 0.2{\scriptsize$\pm$0.1} & 4.7{\scriptsize$\pm$0.5} & 3.2{\scriptsize$\pm$0.4} \\
PruneLLM~\cite{yang2025prunellm} & PruneLLM~\cite{yang2025prunellm} & 43.7{\scriptsize$\pm$0.1} & 47.7{\scriptsize$\pm$0.2} & 46.4{\scriptsize$\pm$0.2} & 6.4{\scriptsize$\pm$0.4} & 9.7{\scriptsize$\pm$0.6} & 8.6{\scriptsize$\pm$0.2} & 0.2{\scriptsize$\pm$0.1} & 0.3{\scriptsize$\pm$0.0} & 0.3{\scriptsize$\pm$0.0} \\
Putri~\cite{coello2026putri} & Putri~\cite{coello2026putri} & 27.1{\scriptsize$\pm$3.4} & 30.1{\scriptsize$\pm$0.9} & 29.1{\scriptsize$\pm$1.7} & 9.5{\scriptsize$\pm$4.1} & 10.0{\scriptsize$\pm$5.6} & 9.8{\scriptsize$\pm$5.1} & -- & -- & -- \\
PORTA~\cite{MinSeok2026PORTA} & PORTA~\cite{MinSeok2026PORTA} & 13.7{\scriptsize$\pm$1.2} & 19.3{\scriptsize$\pm$2.3} & 17.4{\scriptsize$\pm$1.9} & 1.5{\scriptsize$\pm$0.2} & 3.2{\scriptsize$\pm$0.1} & 2.7{\scriptsize$\pm$0.1} & 1.0{\scriptsize$\pm$0.2} & 2.4{\scriptsize$\pm$0.4} & 1.9{\scriptsize$\pm$0.3} \\
TRSP~\cite{feng2026trsp} & TRSP~\cite{feng2026trsp} & 25.7{\scriptsize$\pm$2.4} & 37.0{\scriptsize$\pm$2.2} & 33.2{\scriptsize$\pm$2.2} & 2.1{\scriptsize$\pm$1.0} & 4.9{\scriptsize$\pm$1.6} & 3.9{\scriptsize$\pm$1.4} & 0.4{\scriptsize$\pm$0.4} & 0.9{\scriptsize$\pm$0.9} & 0.7{\scriptsize$\pm$0.7} \\
SVD-LLM~\cite{wang2025svdllm} & SVD-LLM~\cite{wang2025svdllm} & 47.6{\scriptsize$\pm$0.5} & 42.6{\scriptsize$\pm$0.8} & 44.2{\scriptsize$\pm$0.6} & 24.7{\scriptsize$\pm$0.8} & 24.4{\scriptsize$\pm$0.6} & 24.5{\scriptsize$\pm$0.6} & 7.6{\scriptsize$\pm$1.0} & 11.6{\scriptsize$\pm$2.0} & 10.3{\scriptsize$\pm$1.5} \\
AdaSVD~\cite{Li_2026_CVPR_Adasvd} & AdaSVD~\cite{Li_2026_CVPR_Adasvd} & 45.9{\scriptsize$\pm$0.6} & 42.5{\scriptsize$\pm$0.7} & 43.6{\scriptsize$\pm$0.6} & 29.6{\scriptsize$\pm$1.1} & 26.7{\scriptsize$\pm$0.7} & 27.6{\scriptsize$\pm$0.8} & 20.7{\scriptsize$\pm$1.1} & 19.5{\scriptsize$\pm$0.6} & 19.9{\scriptsize$\pm$0.8} \\
\methodabbr (ours) & Uniform w/ \overparamabbr & \textbf{52.2}{\scriptsize$\pm$0.2} & \textbf{52.8}{\scriptsize$\pm$0.3} & \textbf{52.6}{\scriptsize$\pm$0.2} & \textbf{35.8}{\scriptsize$\pm$0.6} & \textbf{32.6}{\scriptsize$\pm$0.5} & \textbf{33.7}{\scriptsize$\pm$0.5} & \textbf{23.6}{\scriptsize$\pm$0.5} & \textbf{22.3}{\scriptsize$\pm$0.2} & \textbf{22.8}{\scriptsize$\pm$0.1} \\
\bottomrule
\end{tabular}
\vspace{-2mm}
\label{tab:strategy}
\end{table*}

\begin{table*}[t]
\centering
\caption{Results on structured weight pruning/\taskabbr results averaged over three runs on an InternVL3-2B~\cite{zhu2025internvl3} backbone after performing recovery via finetuning on the Seen datasets. We find our \methodabbr approach offers 1-3.5\% gains over prior work.  See  \cref{sec:recovery} for discussion on additional recovery strategies.}
\vspace{-2mm}
\begin{tabular}{lcccccc}
\toprule
\flops Budget & \multicolumn{3}{c}{30\%} & \multicolumn{3}{c}{50\%} \\
\cmidrule(lr){2-4} \cmidrule(lr){5-7}
Method & Seen & Unseen & All & Seen & Unseen & All \\
\midrule
Wanda~\cite{sun2024wanda} & 61.4{\scriptsize$\pm$0.7} & 32.1{\scriptsize$\pm$0.2} & 41.9{\scriptsize$\pm$0.3} & 22.6{\scriptsize$\pm$3.1} & 8.5{\scriptsize$\pm$2.8} & 13.2{\scriptsize$\pm$2.9} \\
D$^2$Prune~\cite{xiong2026d2prune} & 53.7{\scriptsize$\pm$2.1} & 28.8{\scriptsize$\pm$0.8} & 37.1{\scriptsize$\pm$1.2} & 43.9{\scriptsize$\pm$0.7} & 23.6{\scriptsize$\pm$0.4} & 30.4{\scriptsize$\pm$0.3} \\ 
PORTA~\cite{MinSeok2026PORTA} & 63.4{\scriptsize$\pm$0.8} & 33.7{\scriptsize$\pm$0.5} & 43.6{\scriptsize$\pm$0.6} & 21.9{\scriptsize$\pm$5.3} & 9.8{\scriptsize$\pm$3.9} & 13.8{\scriptsize$\pm$4.4} \\
RCPU~\cite{haruta2026rcpu} & 63.2{\scriptsize$\pm$0.5} & 34.6{\scriptsize$\pm$0.4} & 44.2{\scriptsize$\pm$0.4} & 43.4{\scriptsize$\pm$0.9} & 23.1{\scriptsize$\pm$1.2} & 29.9{\scriptsize$\pm$1.1} \\
TRSP~\cite{feng2026trsp} & 52.3{\scriptsize$\pm$2.0} & 31.5{\scriptsize$\pm$1.0} & 38.4{\scriptsize$\pm$1.2} & 36.2{\scriptsize$\pm$0.4} & 20.5{\scriptsize$\pm$1.5} & 25.8{\scriptsize$\pm$0.9} \\
SVD-LLM~\cite{wang2025svdllm} & 56.4{\scriptsize$\pm$3.1} & 29.7{\scriptsize$\pm$0.3} & 38.6{\scriptsize$\pm$1.1} & 38.1{\scriptsize$\pm$1.7} & 21.8{\scriptsize$\pm$0.6} & 27.2{\scriptsize$\pm$0.6} \\
AdaSVD~\cite{Li_2026_CVPR_Adasvd} & 60.3{\scriptsize$\pm$1.1} & 31.7{\scriptsize$\pm$0.8} & 41.2{\scriptsize$\pm$0.3} & 46.9{\scriptsize$\pm$0.8} & 24.6{\scriptsize$\pm$0.5} & 32.0{\scriptsize$\pm$0.6} \\
Basis Sharing~\cite{wang2025basissharing} & 53.3{\scriptsize$\pm$1.3} & 29.1{\scriptsize$\pm$0.3} & 37.2{\scriptsize$\pm$0.6} & 33.9{\scriptsize$\pm$1.0} & 20.0{\scriptsize$\pm$1.1} & 24.7{\scriptsize$\pm$0.7} \\ 
Cat~\cite{lu2026skipcat} & 61.5{\scriptsize$\pm$0.8} & 31.4{\scriptsize$\pm$0.6} & 41.5{\scriptsize$\pm$0.7} & 45.9{\scriptsize$\pm$3.0} & 24.9{\scriptsize$\pm$0.3} & 31.9{\scriptsize$\pm$0.8} \\
\methodabbr w/ \overparamabbr (ours) & \textbf{63.9}{\scriptsize$\pm$0.9} & \textbf{35.5}{\scriptsize$\pm$0.4} & \textbf{45.0}{\scriptsize$\pm$0.5} & \textbf{51.8}{\scriptsize$\pm$0.9} & \textbf{27.3}{\scriptsize$\pm$0.4} & \textbf{35.5}{\scriptsize$\pm$0.4} \\
\bottomrule
\end{tabular}
\vspace{-2mm}
\label{tab:recovery}
\end{table*}

\begin{table}[t]
\centering
\caption{Results on criteria-only structured weight pruning/\taskabbr averaged over three runs across five architectures using only 256 samples from CLEVRER~\cite{yi2020clevrer} (one of the three Seen datasets) for any calibration.  For every method, all layers in the network must reduce the same number of \flops (\ie, uniform pruning/\taskabbr).  Our \methodabbr approach demonstrates a strong ability to generalize, reporting the best overall results in 12 of the 15 model and \flops budget settings and taking second best on the remaining 3.}
\vspace{-2mm}
\begin{tabular}{lccc}
\toprule
Method & 10\% All & 30\% All & 50\% All \\
\midrule
\multicolumn{4}{l}{\emph{InternVL3-1B}~\cite{zhu2025internvl3}} \\
SVD-LLM~\cite{wang2025svdllm} & 31.9{\scriptsize$\pm$0.5} & 18.5{\scriptsize$\pm$0.7} & 3.6{\scriptsize$\pm$1.9} \\
Basis Sharing~\cite{wang2025basissharing} & 29.0{\scriptsize$\pm$0.3} & 18.5{\scriptsize$\pm$0.3} & 4.8{\scriptsize$\pm$0.1} \\
Cat~\cite{lu2026skipcat} & 37.2{\scriptsize$\pm$1.6} & 21.1{\scriptsize$\pm$0.7} & 18.2{\scriptsize$\pm$0.3} \\
\methodabbr (ours) & \textbf{38.2}{\scriptsize$\pm$0.5} & \textbf{23.1}{\scriptsize$\pm$0.7} & \textbf{18.5}{\scriptsize$\pm$0.1} \\
\midrule
\multicolumn{4}{l}{\emph{Qwen2-VL-2B}~\cite{wang2024qwen2vl}} \\
SVD-LLM~\cite{wang2025svdllm} & 46.0{\scriptsize$\pm$0.2} & 25.5{\scriptsize$\pm$0.1} & 18.1{\scriptsize$\pm$0.8} \\
Basis Sharing~\cite{wang2025basissharing} & 44.1{\scriptsize$\pm$0.2} & 24.3{\scriptsize$\pm$0.2} & 17.2{\scriptsize$\pm$1.2} \\
Cat~\cite{lu2026skipcat} & 48.7{\scriptsize$\pm$0.3} & 27.0{\scriptsize$\pm$0.4} & 21.1{\scriptsize$\pm$0.2} \\
\methodabbr (ours) & \textbf{50.8}{\scriptsize$\pm$0.2} & \textbf{31.8}{\scriptsize$\pm$0.9} & \textbf{21.8}{\scriptsize$\pm$0.1} \\
\midrule
\multicolumn{4}{l}{\emph{Qwen3-VL-2B}~\cite{bai2025qwen3vl}} \\
SVD-LLM~\cite{wang2025svdllm} & 41.8{\scriptsize$\pm$0.6} & \textbf{25.8}{\scriptsize$\pm$0.2} & 19.9{\scriptsize$\pm$0.6} \\
Basis Sharing~\cite{wang2025basissharing} & 40.4{\scriptsize$\pm$0.7} & 24.5{\scriptsize$\pm$0.6} & 19.4{\scriptsize$\pm$0.4} \\
Cat~\cite{lu2026skipcat} & 41.9{\scriptsize$\pm$0.5} & 24.8{\scriptsize$\pm$0.3} & 20.3{\scriptsize$\pm$0.2} \\
\methodabbr (ours) & \textbf{43.1}{\scriptsize$\pm$0.7} & 25.6{\scriptsize$\pm$0.4} & \textbf{21.0}{\scriptsize$\pm$0.3} \\
\midrule
\multicolumn{4}{l}{\emph{Qwen3.5-2B}~\cite{qwen2026qwen35}} \\
SVD-LLM~\cite{wang2025svdllm} & 34.4{\scriptsize$\pm$0.5} & 24.7{\scriptsize$\pm$0.2} & 18.4{\scriptsize$\pm$0.2} \\
Basis Sharing~\cite{wang2025basissharing} & 33.0{\scriptsize$\pm$0.1} & 24.6{\scriptsize$\pm$0.4} & 18.6{\scriptsize$\pm$0.7} \\
Cat~\cite{lu2026skipcat} & 34.4{\scriptsize$\pm$0.7} & \textbf{25.0}{\scriptsize$\pm$0.3} & 19.3{\scriptsize$\pm$0.0} \\
\methodabbr (ours) & \textbf{34.7}{\scriptsize$\pm$0.6} & 24.8{\scriptsize$\pm$0.4} & \textbf{19.8}{\scriptsize$\pm$0.1} \\
\midrule
\multicolumn{4}{l}{\emph{Gemma 4 E2B}~\cite{gemmateam2026gemma4}} \\
SVD-LLM~\cite{wang2025svdllm} & 32.0{\scriptsize$\pm$0.2} & 19.6{\scriptsize$\pm$0.3} & 14.0{\scriptsize$\pm$0.1} \\
Basis Sharing~\cite{wang2025basissharing} & 29.7{\scriptsize$\pm$0.4} & 18.8{\scriptsize$\pm$0.4} & 13.7{\scriptsize$\pm$0.0} \\
Cat~\cite{lu2026skipcat} & 37.2{\scriptsize$\pm$0.3} & 23.0{\scriptsize$\pm$0.4} & \textbf{16.2}{\scriptsize$\pm$0.1} \\
\methodabbr (ours) & \textbf{38.0}{\scriptsize$\pm$0.2} & \textbf{23.6}{\scriptsize$\pm$0.3} & 15.5{\scriptsize$\pm$0.2} \\
\bottomrule
\end{tabular}
\vspace{-2mm}
\label{tab:architectures}
\end{table}

\noindent\textbf{Datasets.} We evaluate on nine vision-language benchmarks including four with images (GQA~\cite{hudson2019gqa}, VQAv2~\cite{goyal2017vqav2}, VizWiz~\cite{gurari2018vizwiz} and TextVQA~\cite{singh2019textvqa}) and five with videos (CLEVRER~\cite{yi2020clevrer}, NExT-QA~\cite{xiao2021nextqa}, Perception Test~\cite{patraucean2023perception}, VideoMME~\cite{fu2025videomme}, and LongVideoBench~\cite{wu2024longvideobench}). These are split into \textit{Seen} datasets for calibration (256 samples of CLEVRER) or when recovery is employed (134K samples from CLEVRER, Next-QA, and GQA), and \textit{Unseen} evaluation-only datasets (VQAv2, VizWiz, TextVQA, Perception Test, VideoMME, LongVideoBench). See  \cref{sec:datasets} for a detailed discussion.

\noindent\textbf{Metrics.} Multiple-choice questions in NExT-QA, Perception Test, VideoMME, LongVideoBench and parts of CLEVRER report accuracy of correct letter(s). Open-ended GQA and CLEVRER (descriptive) answers require exact match after VQA's answer normalization~\cite{antol2015vqa}. VQAv2, VizWiz and TextVQA use the official soft VQA accuracy, averaged over the ten leave-one-annotator-out subsets, $\frac{1}{10}\sum_i \min(1,\#\{\text{matches among the other nine}\}/3)$. Most tables report the mean accuracy over Seen, Unseen, and over all nine datasets (All).  \cref{sec:perdataset} contains per-dataset results for all main paper tables. We report the mean and standard deviation over three fixed random seeds.

\noindent\textbf{Implementation details.} We tuned any hyperparameters for each method using the Seen validation sets.  Methods that use activation-aware \taskabbr via the calibration split whiten each layer with Cholesky factor of its input second-moment matrix
$S=XX^\top$, regularized by $10^{-4}\cdot\mathrm{mean}(\mathrm{diag}\,S)$. For methods using recovery we used the AdamW optimizer with a batch size of 128 and learning rate $2\times10^{-5}$.  We included a 3\% linear warmup step and used a cosine learning rate scheduler.  Every recovery method was trained for 3 epochs. We evaluated every epoch on the validation split to select the best checkpoint.  On an InternVL3-2B~\cite{zhu2025internvl3} model, most methods took around 2 hours for recovery using $8\times$A100 NVIDIA GPUs. See  \cref{sec:base_arch} for a discussion on backbone selection.

\subsection{Results}

\cref{tab:criteria} reports criteria-only weight pruning/\taskabbr to fit a \flops budget, where every layer in the network is reduced the same amount and only vary on what is retained. \methodabbr gets a 2-3\% overall boost over prior work, with gains on both Seen and Unseen datasets.  Comparing the last three lines of \cref{tab:criteria}, we find that both \balancedabbr (\cref{sec:balanced}) and \downstreamabbr (\cref{sec:downstream}) contribute to their overall performance, with their combinations becoming more effective as \flops decrease.  These results help highlight how effective the weight reduction criteria itself is.  Further,  this setting's uniform shapes suit graph compilers such as XLA in JAX~\cite{frostig2018jax}, NVIDIA's TensorRT~\cite{nvidia_tensorrt}, and \texttt{torch.compile}~\cite{ansel2024pytorch2}, which specialize kernels to static shapes, enabling sharing across identical blocks.  However, many methods in prior work perform very poorly, in part, due to the uniform reduction strategy. In fact, many prior work's (often non-uniform) budget allocation strategy is sometimes listed as a primary contribution (\eg,~\cite{an2024flap,voita2019analyzing,madinei2026interlace,Li_2026_CVPR_Adasvd,wang2025basissharing,lu2026skipcat,yang2025prunellm}).

\cref{tab:strategy} addresses this by exploring different \flops budget allocation strategies.  Some methods focus their \flops reduction entirely in the LLM  (\eg,~\cite{ma2023llmpruner,guo2025slimllm,feng2026trsp,yang2025prunellm}), resulting in higher pruning per-layer to reach \flops budgets that we found can quickly collapse performance.  This could be seen as unfair since other methods may compress both towers.  However, we found prior work often did not generalize beyond their original tower scope as demonstrated by the poor performance of methods like PruneLLM~\cite{yang2025prunellm} in \cref{tab:criteria} and is also echoed in prior work~\cite{TasnimCRISP2026}.

That said, some methods like AdaSVD~\cite{Li_2026_CVPR_Adasvd} proposed a search strategy that \cref{tab:strategy} shows works well factorizing both ViT and LLM towers.  For our own \methodabbr approach that also factorizes both towers, we add extra model capacity without adding test latency via \overparamabbr (\cref{sec:opgroups}), and use uniform \taskabbr for other layers.  Comparing the results from prior work in \cref{tab:strategy} to their corresponding uniform strategy results from \cref{tab:criteria} shows methods improved up to 7\%. However, \cref{tab:strategy} also reports our approach still outperforms them by 3-6\% overall.  Further, many methods we evaluated also create non-uniform block shapes (\eg,~\cite{Li_2026_CVPR_Adasvd}), unlike our approach that still creates uniform sizes that is easily optimized as discussed earlier. See  \cref{sec:opanalysis} for further discussion on \overparamabbr's effect on training.

\cref{tab:recovery} reports performance after a recovery stage where we finetuned all damaged layers using the Seen datasets to help heal affected layers.  Comparing these results to those in the prior two tables, performance jumped up to 41\% across the pruning rates.  \methodabbr also reports gains after recovery of 1-3.5\% over prior work.  However, the benefit from recovery was mostly on Seen datasets, with boosts to Unseen datasets being far smaller.  See  \cref{sec:recovery} for a longer discussion on recovery strategies.

Finally, \cref{tab:architectures} compares \methodabbr to closely related factorization methods over five additional architectures. Our approach gets the best performance in the vast majority of settings, including reporting gains in at least two of three \flops budgets for every model (and is the second best otherwise), demonstrating its ability to generalize.

\subsection{Discussion}

\noindent\textbf{Effect of sharing a basis between layers of different size.} In \cref{sec:balanced} we claimed that using \taskabbr methods of prior work over shared layers of unequal sizes (like the twin layers that both Cat~\cite{lu2026skipcat} and we use) would likely bias the fit towards the larger layer.  \cref{tab:recon_error} demonstrates this occurs in practice by comparing the relative activation-space reconstruction error over the same layers.  We find balancing member contribution can help overall reconstruction.  

That said, unlike similar methods found in prior work, we also consider the possibility that some layers may simply be more important than others. In other words, that some bias might actually be beneficial rather than purely even balancing (\ie, $\gamma \neq 1$ in \cref{eq:balanced}).  \cref{tab:balanced} shows that this can help boost performance over similar methods in prior work.  See  \cref{sec:balanced_study} for a discussion on setting $\gamma$.

\begin{table}[t]
\centering
\setlength{\tabcolsep}{1.pt}
\caption{Comparison of relative activation-space reconstruction error $\|(W-\hat W)X\|_F/\|WX\|_F$ (\%) averaged over three runs using only 256 samples from CLEVRER~\cite{yi2020clevrer} (one of the three Seen datasets) for any calibration.  Our \balancedabbr balances the contribution of each member, resulting in lower overall error in affected layers.}
\vspace{-2mm}
\begin{tabular}{lccccccccc}
\toprule
Method & 10\% Recon.\ $\downarrow$ & 30\% Recon.\ $\downarrow$ & 50\% Recon.\ $\downarrow$\\
\midrule
\multicolumn{2}{l}{\textit{InternVL3-2B}~\cite{zhu2025internvl3}} \\
Cat~\cite{lu2026skipcat} & 2.77{\scriptsize$\pm$0.11}& 4.65{\scriptsize$\pm$0.15} & 7.45{\scriptsize$\pm$0.16} \\
\balancedabbr (\cref{sec:balanced}) & \textbf{2.20}{\scriptsize$\pm$0.09} & \textbf{3.82}{\scriptsize$\pm$0.13} & \textbf{6.37}{\scriptsize$\pm$0.14} \\
\midrule
\multicolumn{2}{l}{\textit{Gemma-4-E2B}~\cite{gemmateam2026gemma4}} \\
Cat~\cite{lu2026skipcat} & 2.41{\scriptsize$\pm$0.06} & 4.19{\scriptsize$\pm$0.08} & 6.59{\scriptsize$\pm$0.10} \\
\balancedabbr (\cref{sec:balanced}) & \textbf{2.13}{\scriptsize$\pm$0.05} & \textbf{3.78}{\scriptsize$\pm$0.08} & \textbf{6.07}{\scriptsize$\pm$0.09} \\
\bottomrule
\end{tabular}
\vspace{-2mm}
\label{tab:recon_error}
\end{table}

\begin{table}[t]
    \centering
    \caption{Comparison on InternVL3-1B~\cite{zhu2025internvl3} replacing \balancedabbr in our full method with BKV balancing used in prior work.  We find \balancedabbr provides a performance boost in most settings.}
    \vspace{-2mm}
    \begin{tabular}{lccc}
    \toprule
    Method & 10\% All & 30\% All & 50\% All \\
    \midrule
    BKV~\cite{meng2025transmla} & 37.5{\scriptsize$\pm$0.7} & \textbf{23.5}{\scriptsize$\pm$0.2} & 17.4{\scriptsize$\pm$0.3} \\
    \balancedabbr (\cref{sec:balanced}) & \textbf{38.2}{\scriptsize$\pm$0.5} & 23.1{\scriptsize$\pm$0.7} & \textbf{18.5}{\scriptsize$\pm$0.1} \\
    \bottomrule
    \end{tabular}
    \vspace{-2mm}
    \label{tab:balanced}
\end{table}

\begin{figure}[t]
  \centering
  \includegraphics[width=\linewidth]{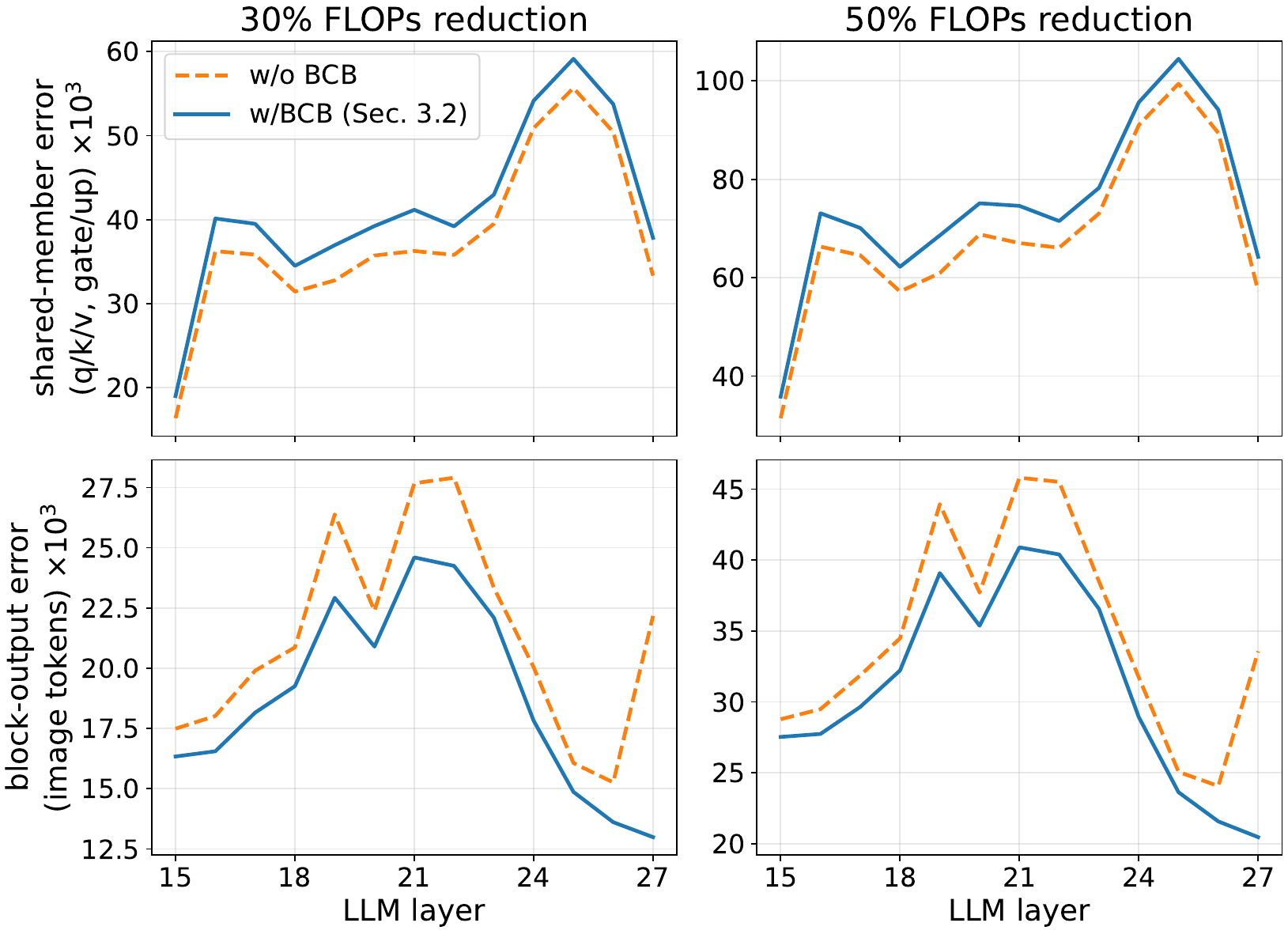}
  \vspace{-6mm}
  \caption{Relative squared error against the dense InternVL3-2B~\cite{zhu2025internvl3} ($\times10^{3}$) in language-model layers 15--27 without recovery, at 30\% (left) and 50\% (right) \flops reduction. \textbf{Top:} reconstruction error of the shared members (query/key/value and gate/up), each evaluated on the dense model's input to it, \ie, the quantity the factorization fits. \textbf{Bottom:} error of each decoder block's output on image tokens, given the dense block input. With \downstreamabbr (\cref{sec:downstream}) the members are reconstructed slightly less accurately, but the error that reaches the block output is consistently lower, especially in the final layer.}
  \vspace{-2mm}
  \label{fig:bcb_contribution}
\end{figure}

\noindent\textbf{Effect of \downstreamabbr on selected factors.} \cref{fig:bcb_contribution} compares the reconstruction quality of \methodabbr with and without \downstreamabbr. In layers 15--27, \downstreamabbr \emph{increases} the shared members' reconstruction error by 8--10\%, yet \emph{decreases} the error of the block output on image tokens by 15--17\%. The row weights $\delta$ in \cref{eq:downstream} redistribute rank toward the rows whose errors propagate to the block output, so a worse fit to $W_k$ yields a better approximation of what the block computes. The effect grows with depth, \eg, the final layer's output error falls by about 40\%, while layers 0--14 change by only 2--3\%. 

\noindent\textbf{Generalization to related settings.}  Some recent work has started to combine different tasks like \taskabbr, weight pruning, quantization, or token pruning to obtain synergistic gains.  We show using \methodabbr as a drop-in replacement boosts performance by 1-9\% at the same \flops budget after combining our \taskabbr method with weight pruning in \cref{tab:plusprune}, quantization in \cref{tab:plusquant}, and token pruning in \cref{tab:plustok}.

\begin{table}[t]
    \centering
    \caption{Evaluates joint \taskabbr and weight pruning on an InternVL3-2B~\cite{zhu2025internvl3} backbone averaged over three runs using only 256 samples from CLEVRER~\cite{yi2020clevrer} (one of the three Seen datasets) for any calibration.  We find that combining \methodabbr w/\overparamabbr and Skip~\cite{lu2026skipcat} weight pruning boosts performance by 1-3.5\%.}
    \vspace{-2mm}
    \begin{tabular}{lccc}
    \toprule
    Method & 10\% All & 30\% All & 50\% All \\
    \midrule
    Cat w/Skip~\cite{lu2026skipcat} & 54.5{\scriptsize$\pm$0.7} & 36.3{\scriptsize$\pm$0.8} & 22.0{\scriptsize$\pm$0.4} \\ 
    \methodabbr (ours) w/ Skip & 55.1{\scriptsize$\pm$0.4} & 39.3{\scriptsize$\pm$0.5}  & 23.3{\scriptsize$\pm$0.3} \\
    \methodabbr w/ \overparamabbr \& Skip & \textbf{55.4}{\scriptsize$\pm$0.2} & \textbf{39.7}{\scriptsize$\pm$0.6}  & \textbf{23.8}{\scriptsize$\pm$0.1} \\
    \bottomrule
    \end{tabular}
    \vspace{-2mm}
    \label{tab:plusprune}
\end{table}

\begin{table}[t]
\centering
\setlength{\tabcolsep}{3pt}
\caption{Results on joint \taskabbr and 4 bit quantization on an InternVL3-2B~\cite{zhu2025internvl3} backbone averaged over three runs after performing recovery using the Seen datasets.  Combining \methodabbr with prior work boosts performance by 3.5-9\%.}
\vspace{-2mm}
\begin{tabular}{lcc}
    \toprule
    Method & 50\% All & 70\% All \\
    \midrule
    GETA~\cite{qu2025geta} & 31.4{\scriptsize$\pm$0.5} & 22.0{\scriptsize$\pm$3.2} \\
    UniQL~\cite{chiang2026uniql} & 33.2{\scriptsize$\pm$1.6} & 22.4{\scriptsize$\pm$1.5} \\
    \methodabbr (ours)  w/ GETA & \textbf{36.5}{\scriptsize$\pm$0.6} & \textbf{31.4}{\scriptsize$\pm$0.6} \\
\bottomrule
\end{tabular}
\vspace{-2mm}
\label{tab:plusquant}
\end{table}

\begin{table}[t]
\centering
\setlength{\tabcolsep}{3pt}
\caption{Results on joint \taskabbr and token pruning on an InternVL3-2B~\cite{zhu2025internvl3} backbone averaged over three runs using only 256 samples from CLEVRER~\cite{yi2020clevrer} (one of the three Seen datasets) for any calibration.  Combining \methodabbr with prior work reports a 2-9\%, even after improving prior methods (see  \cref{sec:baseline}).}
\vspace{-2mm}
\begin{tabular}{lcc}
    \toprule
    Method & 50\% All & 70\% All \\
\midrule
CoMP~\cite{wu2026comp} & 18.0{\scriptsize$\pm$3.5} & 8.5{\scriptsize$\pm$2.6} \\
CoMP w/NLL & 23.1{\scriptsize$\pm$2.1} & 10.8{\scriptsize$\pm$2.8} \\
\methodabbr (ours)  w/ CoMP & \textbf{25.1}{\scriptsize$\pm$0.4} & \textbf{19.6}{\scriptsize$\pm$0.1} \\
\bottomrule
\end{tabular}
\label{tab:plustok}
\end{table}

\noindent\textbf{Latency and memory costs.} \cref{tab:latency} compares the latency and memory costs of competing methods.  As expected, \taskabbr methods like ours do not obtain the full \flops savings due to the overhead from essentially replacing every factorized layer's weights with multiple smaller ones. However, among factorized methods (starting on the third row), ours is the fastest while outperforming prior work by up to 2\%.  That said, Basis Sharing requires less memory.  Also, both weight pruning methods (first two rows) are faster at the same \flops.   Thus, the last row in \cref{tab:latency} reduces \flops further to achieve a 5.5\% gain in task performance while requiring 9\% less memory than Basis Sharing and similar latencies to weight pruning methods. See  \cref{sec:latency} for additional architectures and \flops rates.

\begin{table}[t]
    \centering
    \small
    \setlength{\tabcolsep}{2pt}
    \caption{Compares latency and memory costs with a NVIDIA L40S GPU on an InternVL3-2B~\cite{zhu2025internvl3} backbone averaged over three runs after performing recovery using the Seen datasets. We find \methodabbr obtains a 2\% task performance gain at similar or better latencies and higher memory savings than prior work.}
    \vspace{-2mm}
    \begin{tabular}{lcccc}
    \toprule
     &  & Memory & Latency & Speedup\\
    Method &  50\% All & (GB) &  (ms) & vs.\ Dense\\
    \midrule
    D$^2$Prune~\cite{xiong2026d2prune} & 30.4{\scriptsize$\pm$0.3} & 2.89 & 88.5{\scriptsize$\pm$0.3} & 1.53$\times$ \\
    RCPU~\cite{haruta2026rcpu} & 29.9{\scriptsize$\pm$1.1} & 2.89 & 88.5{\scriptsize$\pm$0.5} & 1.53$\times$ \\
    AdaSVD~\cite{Li_2026_CVPR_Adasvd} & 32.0{\scriptsize$\pm$0.6} & 2.98 & 103.8{\scriptsize$\pm$0.4} & 1.30$\times$ \\
    Basis Sharing~\cite{wang2025basissharing} & 24.7{\scriptsize$\pm$0.7} & 2.84 & 103.8{\scriptsize$\pm$0.1} & 1.30$\times$ \\
    Cat~\cite{lu2026skipcat} & 31.9{\scriptsize$\pm$0.8} & 2.86 & 102.3{\scriptsize$\pm$0.3} & 1.32$\times$ \\
    Cat w/ Skip~\cite{lu2026skipcat} & 37.2{\scriptsize$\pm$0.2} & 2.99 & 106.6{\scriptsize$\pm$0.4} & 1.27$\times$ \\
    \methodabbr w/ \overparamabbr (ours) & 35.5{\scriptsize$\pm$0.4} & 2.88 & 100.8{\scriptsize$\pm$0.9} & 1.34$\times$ \\
    \hspace{2mm} \& Skip & 39.1{\scriptsize$\pm$0.2} & 2.92 & 101.5{\scriptsize$\pm$0.8} & 1.33$\times$ \\
    \hspace{2mm} \& Skip @60\% & 36.0{\scriptsize$\pm$0.3} & 2.58 & 91.2{\scriptsize$\pm$0.7} & 1.48$\times$ \\
    \bottomrule
    \end{tabular}
    \vspace{-2mm}
    \label{tab:latency}
\end{table}

%% file: sec/5_conclusion.tex
\section{Conclusion}

This paper introduced \method (\methodabbr), a shared-basis \taskabbr method that computes a single basis once for each group of twin layers, so more of the \flops it removes translate into lower latency. \methodabbr uses \balancedabbr to remove a bias suffered by prior work in favor of larger group members, and \downstreamabbr keeps factors that matter most for each block's output. Further, we show \overparamabbr can add capacity to rank-sensitive layers during training, which are folded back to their dense size for inference at no extra cost. Across six MLLMs and nine image and video benchmarks, \methodabbr outperforms the state of the art by 2-6\% without finetuning and by 1-3.5\% after recovery at matched \flops, while being the fastest of the factorization methods we evaluated. When used as a drop-in replacement, \methodabbr also improves combinations with weight pruning, quantization, and token pruning by 1-9\%.  Future work could explore mechanisms to leverage more of the \overparamabbr capacity, or more fully explore combinations with related tasks for further gains.

%% file: sec/X_suppl.tex
\clearpage
\appendix

\begin{table*}[tb]
\centering
\setlength{\tabcolsep}{4pt}
\caption{Comparison of \downstreamabbr with the row importance of FWSVD~\cite{hsu2022fwsvd} and MoDeGPT~\cite{lin2025modegpt} on InternVL3-2B~\cite{zhu2025internvl3} averaged over three runs using only 256 samples from CLEVRER~\cite{yi2020clevrer} (one of the Seen datasets) for calibration. Each row replaces \downstreamabbr's row weights $D_k$ in \cref{eq:downstream} on the same shared groups.  We find our \downstreamabbr approach outperforms prior work.  See \cref{sec:downstream_study} for discussion.}
\begin{tabular}{l|ccc|ccc|ccc}
\toprule
\flops Budget & \multicolumn{3}{c|}{10\%} & \multicolumn{3}{c|}{30\%} & \multicolumn{3}{c}{50\%} \\
Row weights & Seen & Unseen & All & Seen & Unseen & All & Seen & Unseen & All \\
\midrule
\methodabbr w/o \downstreamabbr & 48.3{\scriptsize$\pm$0.2} & 46.3{\scriptsize$\pm$0.9} & 47.0{\scriptsize$\pm$0.6} & 33.6{\scriptsize$\pm$0.4} & 29.6{\scriptsize$\pm$0.9} & 30.9{\scriptsize$\pm$0.6} & 23.4{\scriptsize$\pm$0.6} & 20.7{\scriptsize$\pm$0.5} & 21.6{\scriptsize$\pm$0.4} \\
\hspace{3mm} w/ FWSVD~\cite{hsu2022fwsvd} & 48.2{\scriptsize$\pm$0.2} & 46.3{\scriptsize$\pm$0.3} & 46.9{\scriptsize$\pm$0.2} & 33.1{\scriptsize$\pm$0.8} & 28.7{\scriptsize$\pm$0.6} & 30.1{\scriptsize$\pm$0.5} & 23.2{\scriptsize$\pm$0.1} & \textbf{21.6}{\scriptsize$\pm$0.3} & 22.1{\scriptsize$\pm$0.2} \\
\hspace{3mm} w/ MoDeGPT~\cite{lin2025modegpt} & 48.3{\scriptsize$\pm$0.4} & 46.4{\scriptsize$\pm$1.0} & 47.0{\scriptsize$\pm$0.6} & 33.8{\scriptsize$\pm$0.8} & 30.3{\scriptsize$\pm$1.4} & 31.4{\scriptsize$\pm$1.1} & \textbf{23.5}{\scriptsize$\pm$0.6} & 20.7{\scriptsize$\pm$1.0} & 21.7{\scriptsize$\pm$0.9} \\
\methodabbr (ours) & \textbf{48.7}{\scriptsize$\pm$0.2} & \textbf{46.9}{\scriptsize$\pm$0.5} & \textbf{47.5}{\scriptsize$\pm$0.3} & \textbf{34.4}{\scriptsize$\pm$1.0} & \textbf{31.2}{\scriptsize$\pm$0.9} & \textbf{32.2}{\scriptsize$\pm$0.9} & 23.3{\scriptsize$\pm$0.7} & \textbf{21.6}{\scriptsize$\pm$0.5} & \textbf{22.2}{\scriptsize$\pm$0.5} \\
\bottomrule
\end{tabular}
\label{tab:bcb_compare}
\end{table*}

\section{\downstreamabbr Module Discussion}

Below we expand upon on discussion in the main paper to give examples of how we compute our factor weights $\delta$ in \cref{sec:rowweights} and provide a direct comparison to similar ideas from prior work in \cref{sec:downstream_study}.

\subsection{Computing Row Weights $\delta$}
\label{sec:rowweights}

\cref{sec:downstream} we describe how we use $\delta$ to weight potential factors by how informative they are to the block output.  Specifically, \cref{eq:downstream} uses first order estimates of row weights $\delta_k$ that indicate how strongly an error in each output row of member $W_k$ propagates to the output of the block that contains $\mathcal{W}$. Throughout, $x_t$ denotes calibration token $t$ (a column of $X$),
$\mathbb{E}_t[\cdot]$ the average over the calibration set, and
$\operatorname{rms}_t(a)=(\mathbb{E}_t[a_t^2])^{1/2}$. All statistics are measured once on the
\emph{dense} model, before any layer is factorized. For a member row $i$ with error $e_{k,i,t}$
on token $t$, we write the induced change of the block output as
$\Delta y_t\approx g_{k,i,t}\,e_{k,i,t}$ and set
\begin{equation}
  \label{eq:rowweight}
  \delta_{k,i} \;=\; \big(\mathbb{E}_t\,\|g_{k,i,t}\|_2^2\big)^{1/2},
\end{equation}
\ie, the calibration average of the row's sensitivity that replaces the per-token sensitivity in
\cref{eq:downstream}. Cross-row terms are dropped, as in \cref{sec:downstream}. Below we derive
$g_{k,i,t}$ for the two groups that share an input in the models we study. Singletons
(\eg, the attention output projection, the MLP down projection and every vision-tower linear,
including its fused query/key/value projection) use $\delta_k=\mathbf{1}$, for which
\cref{eq:downstream} reduces to \cref{eq:balanced}.

\subsubsection{Gated MLP (gate/up group)}
A gated MLP computes $y=W_{\mathrm{d}}\big(\sigma(g)\odot u\big)$ with $g=W_{\mathrm{g}}x$,
$u=W_{\mathrm{u}}x$ and activation $\sigma$ (SiLU for SwiGLU~\cite{shazeer2020glu}). An error $e$
in channel $c$ of the gate or the up projection changes the output by
\begin{align}
  \Delta y^{(\mathrm{g})}_t &= W_{\mathrm{d}}[:,c]\;\sigma'(g_{c,t})\,u_{c,t}\;e, \nonumber\\
  \Delta y^{(\mathrm{u})}_t &= W_{\mathrm{d}}[:,c]\;\sigma(g_{c,t})\;e,
\end{align}
so by \cref{eq:rowweight}:
\begin{align}
  \label{eq:rw-mlp}
  \delta_{\mathrm{g},c} &= \|W_{\mathrm{d}}[:,c]\|_2\;\operatorname{rms}_t\!\big(\sigma'(g_{c,t})\,u_{c,t}\big), \nonumber\\
  \delta_{\mathrm{u},c} &= \|W_{\mathrm{d}}[:,c]\|_2\;\operatorname{rms}_t\!\big(\sigma(g_{c,t})\big).
\end{align}
A channel whose down-projection column is small, or whose gate is rarely open, receives little weight.  This signal is analogous to some weight pruning methods that use activations to decide what weight to remove (\eg,~\cite{sun2024wanda,an2024flap}), but extended to help decide what factors are worth retaining.

\subsubsection{Grouped-query attention (query/key/value group)}
Let head $h$ use key/value group $\kappa(h)$, with head dimension $d_h$, and let the attention
logits be $s_{h,ij}=q_{h,i}^\top k_{\kappa(h),j}/\sqrt{d_h}$. We ignore the softmax Jacobian and
measure query and key errors in logit space and value errors in output space:
\begin{itemize}
  \item \emph{Query row $(h,d)$.} An error $e$ in $q_{h,i,d}$ changes the logits by
        $\Delta s_{h,ij}=e\,k_{\kappa(h),j,d}/\sqrt{d_h}$ for every key $j$, so
        $\delta_{\mathrm{q},(h,d)}=\operatorname{rms}_t\big(k_{\kappa(h),t,d}\big)$.
  \item \emph{Key row $(\kappa,d)$.} An error in $k_{\kappa,j,d}$ reaches the logits of every
        query head $h$ that reads group $\kappa$, so
        $\delta_{\mathrm{k},(\kappa,d)}=\big(\sum_{h:\kappa(h)=\kappa}\mathbb{E}_t[q_{h,t,d}^2]\big)^{1/2}$.
  \item \emph{Value row $(\kappa,d)$.} An error in $v_{\kappa,j,d}$ is mixed by the attention
        weights (whose rows sum to one) into slot $(h,d)$ of every head $h$ in group $\kappa$,
        and then mapped by the output projection $W_{\mathrm{o}}$, so
        $\delta_{\mathrm{v},(\kappa,d)}=\big(\sum_{h:\kappa(h)=\kappa}\|W_{\mathrm{o}}[:,\,h d_h+d]\|_2^2\big)^{1/2}$.
\end{itemize}
The constant $1/\sqrt{d_h}$ is dropped since it scales a whole member. With rotary position
embeddings~\cite{su2024roformer}, $q$ and $k$ are rotated pairwise by a position-dependent angle,
which preserves each pair's energy, but mixes its two coordinates. Thus, we replace each second
moment by its average over the rotated pair $(d,\,d+d_{\mathrm{rot}}/2)$, so the statistics of
the projection outputs (before rotation) are valid after it. Dimensions outside the rotary span
keep their own moments.

One issue with the approach described so far is that it introduces an inconsistency.  Specifically, query and key errors are measured in logit space, but value errors are measured in output space.  Thus, we normalize each attention
member's weights to unit mean square,
$\delta_k\leftarrow\delta_k/\big(\tfrac{1}{m_k}\|\delta_k\|_2^2\big)^{1/2}$, so $D_k$ only
redistributes rank among the rows within a member.  We set the balance between members via \balancedabbr's
$\alpha_k$ (see \cref{sec:balanced}).
\smallskip

\noindent\textbf{Architecture variants.}
Two variants in our model set need small adaptations. \emph{Gated attention} (Qwen3.5~\cite{qwen2026qwen35}) emits a
per-head output gate alongside each query head. The gate rows have no logit-space term, so they
receive the root-mean-square of that head's query weights, which leaves the gate rows at the
member's average weight. \emph{Layers without a full query/key/value group} (\eg, Gemma~4's~\cite{gemmateam2026gemma4}
has layers that reuse another layer's keys and values) have no group to weight, and their present
projections use $\delta_k=\mathbf{1}$.

\subsection{Comparison to Prior Work}
\label{sec:downstream_study}

Below we describe how we adapt methods from prior work to act as the $\delta$ weights for \cref{eq:downstream}.
\smallskip

\noindent\textbf{FWSVD~\cite{hsu2022fwsvd}.} This method weights each output row of a matrix by its Fisher information, \ie, the summed squared gradients of the task loss with respect to that row's weights, before the SVD. Thus, rows the loss is more sensitive to are reconstructed more accurately, trading a worse weight reconstruction for task accuracy closer to the original model. \emph{Implementation details:} gradients are taken per sample on the dense model with our calibration language-modeling loss over the 256 CLEVRER calibration samples. Each member's weights are normalized to unit mean square, so they only redistribute rank within a member and \downstreamabbr sets the balance between members.
\smallskip

\noindent\textbf{MoDeGPT~\cite{lin2025modegpt}.} This approach compresses pairs of consecutive matrices so that their joint output is preserved, using a Nystr\"om approximation that keeps MLP channels with the highest ridge leverage scores of the intermediate activations, a CR decomposition that keeps query/key dimensions with the largest product of query and key activation norms, and an SVD of the activation-weighted value/output product. \emph{Implementation details:} we use MoDeGPT's selection scores as the row weights $\delta$ in \cref{eq:downstream}, with each row's squared weight proportional to its score. Gate and up rows share the ridge leverage of their channel ($\lambda{=}1$, as in their paper), query rows use the per-dimension query/key norm product, and key rows use the paper's grouped-query group score. The value/output SVD has no per-row score, so value rows are weighted by the norm of the output-projection columns they feed, which is similar to \downstreamabbr's value weight. In addition, as with \downstreamabbr, the rotated query/key pairs share their second moments, and each member is normalized to unit mean square.

\subsubsection{Results}

\cref{tab:bcb_compare} replaces \downstreamabbr's row weights with those of FWSVD~\cite{hsu2022fwsvd} and MoDeGPT~\cite{lin2025modegpt}, keeping every other part of \methodabbr fixed. \downstreamabbr obtains the best overall result at every budget, with the clearest gain at 30\% where it reports a 1-2\% gain overall.  Notably, neither alternative consistently improves over using no row weights at all. FWSVD lowers All at 10\% and 30\% and raises it only at 50\%, which suggests that the gradient of a calibration language-modeling loss is a noisy guide to which rows matter. MoDeGPT's ridge leverage on gate/up members measures how much a channel contributes to the span of the intermediate activations, while \downstreamabbr also accounts for how strongly the channel reaches the block output through the down projection and the gate. Overall, these results suggest block-output sensitivity is a more useful signal than similar local measures (MoDeGPT) or a focus on the final output (FWSVD).

\section{Effect of \overparamabbr (\cref{sec:opgroups}) on Training}
\label{sec:opanalysis}

\begin{figure*}[t]
\centering
\includegraphics[width=0.8\textwidth]{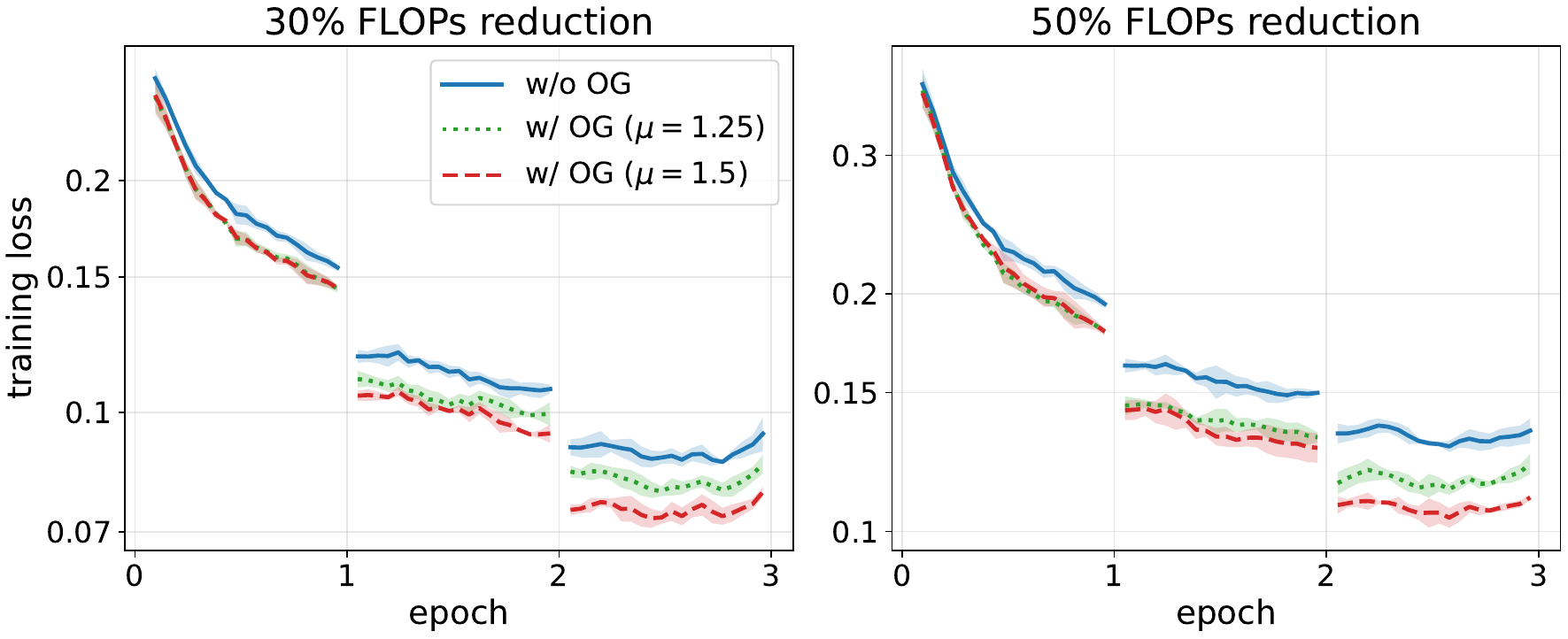}
\caption{Recovery training loss on InternVL3-2B~\cite{zhu2025internvl3} without \overparamabbr and with the vision MLP held at dense cost (\cref{eq:overparam}) at $\mu{=}\mu^\ast{=}1.25$, the smallest rank that reproduces the dense layers exactly (\cref{eq:overparam-exact}), and at $\mu{=}1.5$. Both \overparamabbr variants have the same inference \flops and the same model before recovery, \ie, they differ only in the extra rank directions available during training. Lines show the mean over 3 seeds and bands $\pm$1 standard deviation (5-point moving average within each epoch). The loss steps down at each epoch boundary because recovery repeats the training data, suggesting that some training samples have been memorized. See \cref{sec:opanalysis} for discussion.}
\label{fig:og_loss}
\end{figure*}

\overparamabbr makes two notable changes when it is applied.  First, it keeps the chosen groups at dense cost, which forces the rest of the model to be compressed harder to meet the same budget.  Second, groups where $\mu>\mu^\ast$ have more factor ranks than they need to represent their dense layers. To separate the two, \cref{fig:og_loss} compares the vision MLP at $\mu{=}\mu^\ast{=}1.25$ and $\mu{=}1.5$ against no \overparamabbr at the 30\% and 50\% budgets. At $\mu^\ast$ the factors can represent the dense layers exactly but have no spare rank.  Thus, the two \overparamabbr variants share the same inference \flops, the same compression rate of every other layer, and the same model before recovery (their no-recovery accuracy agrees within 0.5\%). Therefore, any difference after recovery comes from the extra rank directions that are trained like the remaining factors.

\cref{fig:og_loss} shows that keeping the vision MLP at dense cost accounts for some of the drop in training loss. In the third epoch, $\mu{=}1.25$ lowers the loss from 0.089 to 0.082 at 30\% and from 0.132 to 0.117 at 50\%, and the gap is already present in the first epoch (0.190 vs.\ 0.178 and 0.244 vs.\ 0.229). The extra rank at $\mu{=}1.5$ lowers it further, to 0.075 and 0.107, respectively, but this benefit only appears in the later epochs. Since the function class of the two variants is identical, the extra rank changes how the factors are optimized rather than what they can represent.  This is consistent with the implicit acceleration observed when linear layers are overparameterized as products of factors~\cite{arora2018optimization}. The extra directions also stay small: after recovery, their rank-one terms $\|u_i\|\|v_i\|$ are about 1--2\% of the largest direction.  Future work could alter the recovery strategy to make better use of these new factors.

\section{Analyzing Recovery Strategies}
\label{sec:recovery}

We compare different recovery strategies on the same pruning/\taskabbr method to evaluate how it might affect final performance. Every recovery strategy in our experiments receive the same training budget.  Specifically, the \flops of three epochs of our reference SFT (fine-tuning every compressed module of both towers) of \methodabbr at that budget, measured per training sample. For example, LoRA's~\cite{hu2022lora} cheaper backward pass buys about 1 additional epoch. All strategies use the Seen training mix, and the final step picks the best epoch on validation data.  Below we list the different strategies we tested sampled from prior work.
\begin{itemize}
\item \textbf{LoRA}~\cite{hu2022lora} adapters with $r{=}8$, $\alpha{=}16$, dropout $0.05$ on every block linear of both towers, learning rate $10^{-4}$ trained for around 4.1-4.2 epochs. \methodabbr adapters are on both factors, including the shared bases.
\item \textbf{Staged LoRA}~\cite{wang2025svdllm} (\methodabbr only) uses LoRA on the output factors for half the recovery budget, then on the input factors for the other half (about 2.1 epochs each).
\item \textbf{Full Supervised Fine-Tuning (SFT)} trains every parameter, including the embeddings, vision projector, LM head, and \methodabbr's shared bases, with learning rate $2\times10^{-5}$ (3 epochs).  This is the default used in \cref{tab:recovery}.
\item \textbf{Last 3 Blocks}~\cite{lu2026reassessing} trains only the last three LLM blocks and the LM head (about 5.8--6.5 epochs).
\item \textbf{KD}~\cite{hinton2015distilling} uses KL divergence for recovery, \ie, an objective of $(1-\alpha)\,\mathcal{L}_{\mathrm{CE}} + \alpha\,\mathcal{L}_{\mathrm{KL}}$ against the dense InternVL3-2B's logits, with $\alpha{=}0.25$ and temperature $T{=}2$ (results in about 1.8--2.2 epochs, since each step also has a teacher).
\item \textbf{Block Distill.\ $\to$ SFT}~\cite{xu2024besa,chen2025streamline,guo2024ebft} splits the recovery budget into two parts.  The first half is spent on block-wise distillation, where every block of both towers receives the dense model's input to that block and is trained to match the dense block's output under a relative MSE. The second half is our reference SFT (1.5 epochs).
\item \textbf{Progressive (3 Stages)}~\cite{han2015learning} compresses the model in stages to achieve $x/3$, $2x/3$, then $x$ of the target budget using one third of the target budget (one epoch) of SFT after each stage. For \methodabbr, each stage folds the recovered factors back to dense weights and re-factorizes them with Gram statistics measured on the current model. For PORTA, each stage re-prunes the recovered model to $k/3$ of the final per-layer counts from PORTA's allocation and composes the result with the previous stage's pruning.
\end{itemize}

We also report performance that pairs a source method with its own recovery strategy to help alleviate a concern that some methods may simply be well suited to a particular recovery method.  This includes: Minitron~\cite{muralidharan2024minitron} (activation-based width importance, pure logit KD), RLP~\cite{lu2026reassessing} (removing the last blocks, then fine-tuning the last three blocks and the LM head), INTERLACE~\cite{madinei2026interlace} (removing consecutive layer triplets and fine-tuning their neighbors; it cannot reach 50\% because it removes at most $\lfloor L/3 \rfloor$ layers), and Structured GMP~\cite{zhu2018prune} (structured magnitude pruning from the dense model under a cubic sparsity schedule).

\begin{table*}[t]
\centering
\small
\setlength{\tabcolsep}{2pt}
\caption{Results of recovery strategies averaged over three runs on an InternVL3-2B~\cite{zhu2025internvl3} backbone  evaluated on the first 500 test samples of each dataset after recovery with the same budget using the Seen datasets. INTERLACE cannot reach 50\% as it removes at most $\lfloor L/3\rfloor$ layers. Best results are bolded, second best are underlined.  Note that \methodabbr does not use \overparamabbr in these experiments, refer to \cref{tab:recovery} for performance with \overparamabbr.  See \cref{sec:recovery} for further discussion.}
\begin{tabular}{llccccccccc}
\toprule
\flops Budget & & \multicolumn{3}{c}{10\%} & \multicolumn{3}{c}{30\%} & \multicolumn{3}{c}{50\%} \\
\cmidrule(lr){3-5} \cmidrule(lr){6-8} \cmidrule(lr){9-11}
Method & Recovery Strategy & Seen & Unseen & All & Seen & Unseen & All & Seen & Unseen & All \\
\midrule
\multirow{7}{*}{\methodabbr (ours)} & LoRA & 70.1{\scriptsize$\pm$0.4} & 46.2{\scriptsize$\pm$2.3} & 54.2{\scriptsize$\pm$1.4} & \underline{62.5}{\scriptsize$\pm$0.7} & 33.4{\scriptsize$\pm$0.8} & 43.1{\scriptsize$\pm$0.8} & \underline{53.8}{\scriptsize$\pm$0.1} & 27.0{\scriptsize$\pm$0.5} & \underline{36.0}{\scriptsize$\pm$0.3} \\
 & Staged LoRA & 68.4{\scriptsize$\pm$1.0} & 48.4{\scriptsize$\pm$0.8} & 55.1{\scriptsize$\pm$0.4} & 60.3{\scriptsize$\pm$0.5} & 32.3{\scriptsize$\pm$0.5} & 41.6{\scriptsize$\pm$0.5} & 48.4{\scriptsize$\pm$1.6} & 27.0{\scriptsize$\pm$0.3} & 34.2{\scriptsize$\pm$0.4} \\
 & SFT & 70.5{\scriptsize$\pm$0.6} & 48.7{\scriptsize$\pm$0.5} & 56.0{\scriptsize$\pm$0.5} & 61.9{\scriptsize$\pm$1.9} & 35.2{\scriptsize$\pm$0.9} & \underline{44.1}{\scriptsize$\pm$0.2} & 50.1{\scriptsize$\pm$0.6} & 27.1{\scriptsize$\pm$0.3} & 34.8{\scriptsize$\pm$0.2} \\
 & Last 3 Blocks & 50.2{\scriptsize$\pm$1.3} & 42.9{\scriptsize$\pm$2.8} & 45.3{\scriptsize$\pm$2.0} & 36.1{\scriptsize$\pm$2.0} & 29.4{\scriptsize$\pm$1.1} & 31.6{\scriptsize$\pm$0.4} & 29.3{\scriptsize$\pm$1.1} & 24.4{\scriptsize$\pm$2.0} & 26.0{\scriptsize$\pm$1.7} \\
 & KD & 65.4{\scriptsize$\pm$0.6} & 46.4{\scriptsize$\pm$0.7} & 52.7{\scriptsize$\pm$0.5} & 57.5{\scriptsize$\pm$0.9} & 34.6{\scriptsize$\pm$0.5} & 42.2{\scriptsize$\pm$0.6} & 47.2{\scriptsize$\pm$0.3} & \textbf{28.1}{\scriptsize$\pm$0.1} & 34.4{\scriptsize$\pm$0.1} \\
 & Block Distill. $\to$ SFT & 70.2{\scriptsize$\pm$0.7} & 49.0{\scriptsize$\pm$1.6} & \underline{56.1}{\scriptsize$\pm$1.3} & 62.3{\scriptsize$\pm$1.6} & 34.0{\scriptsize$\pm$0.5} & 43.4{\scriptsize$\pm$0.7} & 50.1{\scriptsize$\pm$1.1} & \underline{27.6}{\scriptsize$\pm$0.6} & 35.1{\scriptsize$\pm$0.7} \\
 & Progressive (3 Stages) & 69.0{\scriptsize$\pm$0.2} & 44.5{\scriptsize$\pm$0.4} & 52.7{\scriptsize$\pm$0.2} & 62.4{\scriptsize$\pm$1.6} & 34.6{\scriptsize$\pm$0.8} & 43.9{\scriptsize$\pm$0.7} & \textbf{54.8}{\scriptsize$\pm$1.5} & 27.0{\scriptsize$\pm$0.6} & \textbf{36.3}{\scriptsize$\pm$0.9} \\
\midrule
\multirow{6}{*}{PORTA~\cite{MinSeok2026PORTA}} & LoRA & \underline{71.6}{\scriptsize$\pm$1.1} & 46.2{\scriptsize$\pm$3.4} & 54.6{\scriptsize$\pm$1.9} & \textbf{64.4}{\scriptsize$\pm$0.5} & 34.5{\scriptsize$\pm$1.3} & \textbf{44.4}{\scriptsize$\pm$1.1} & 29.4{\scriptsize$\pm$8.1} & 11.3{\scriptsize$\pm$4.0} & 17.3{\scriptsize$\pm$5.3} \\
 & SFT & \textbf{71.8}{\scriptsize$\pm$0.5} & 46.9{\scriptsize$\pm$1.5} & 55.2{\scriptsize$\pm$1.1} & 62.1{\scriptsize$\pm$0.5} & 34.2{\scriptsize$\pm$1.2} & 43.5{\scriptsize$\pm$0.9} & 24.8{\scriptsize$\pm$1.5} & 13.7{\scriptsize$\pm$1.4} & 17.4{\scriptsize$\pm$1.3} \\
 & Last 3 Blocks & 31.4{\scriptsize$\pm$2.0} & 34.0{\scriptsize$\pm$2.4} & 33.2{\scriptsize$\pm$2.2} & 20.6{\scriptsize$\pm$1.4} & 12.7{\scriptsize$\pm$2.9} & 15.3{\scriptsize$\pm$2.2} & 21.5{\scriptsize$\pm$0.4} & 14.2{\scriptsize$\pm$1.2} & 16.6{\scriptsize$\pm$0.7} \\
 & KD & 68.0{\scriptsize$\pm$1.3} & 49.5{\scriptsize$\pm$0.3} & 55.7{\scriptsize$\pm$0.3} & 57.6{\scriptsize$\pm$1.4} & \underline{35.5}{\scriptsize$\pm$0.7} & 42.9{\scriptsize$\pm$0.1} & 22.2{\scriptsize$\pm$3.0} & 16.6{\scriptsize$\pm$0.6} & 18.5{\scriptsize$\pm$1.4} \\
 & Block Distill. $\to$ SFT & \textbf{71.8}{\scriptsize$\pm$0.4} & \underline{51.9}{\scriptsize$\pm$0.8} & \textbf{58.6}{\scriptsize$\pm$0.5} & 59.6{\scriptsize$\pm$1.3} & \underline{35.5}{\scriptsize$\pm$0.1} & 43.5{\scriptsize$\pm$0.4} & 43.2{\scriptsize$\pm$4.1} & 22.0{\scriptsize$\pm$1.8} & 29.0{\scriptsize$\pm$2.5} \\
 & Progressive (3 Stages) & 71.0{\scriptsize$\pm$0.4} & 47.2{\scriptsize$\pm$0.9} & 55.1{\scriptsize$\pm$0.6} & 60.6{\scriptsize$\pm$0.1} & 32.5{\scriptsize$\pm$0.8} & 41.9{\scriptsize$\pm$0.5} & 46.6{\scriptsize$\pm$1.7} & 23.3{\scriptsize$\pm$1.3} & 31.1{\scriptsize$\pm$0.6} \\
\midrule
Minitron~\cite{muralidharan2024minitron} & Knowledge Distillation & 53.2{\scriptsize$\pm$0.4} & \textbf{52.7}{\scriptsize$\pm$0.1} & 52.9{\scriptsize$\pm$0.1} & 42.5{\scriptsize$\pm$1.2} & \textbf{38.2}{\scriptsize$\pm$2.6} & 39.6{\scriptsize$\pm$1.9} & 20.1{\scriptsize$\pm$3.0} & 16.8{\scriptsize$\pm$2.4} & 17.9{\scriptsize$\pm$2.5} \\
RLP~\cite{lu2026reassessing} & Layer Reverse Order & 58.9{\scriptsize$\pm$0.4} & 50.6{\scriptsize$\pm$0.1} & 53.3{\scriptsize$\pm$0.1} & 39.1{\scriptsize$\pm$1.0} & 24.3{\scriptsize$\pm$0.6} & 29.2{\scriptsize$\pm$0.5} & 8.4{\scriptsize$\pm$0.3} & 10.2{\scriptsize$\pm$0.3} & 9.6{\scriptsize$\pm$0.2} \\
INTERLACE~\cite{madinei2026interlace} & Consecutive Layers & 55.4{\scriptsize$\pm$0.4} & 49.6{\scriptsize$\pm$0.6} & 51.5{\scriptsize$\pm$0.4} & 30.2{\scriptsize$\pm$1.6} & 21.6{\scriptsize$\pm$1.2} & 24.5{\scriptsize$\pm$1.3} & -- & -- & -- \\
Structured GMP~\cite{zhu2018prune} & Gradual Sparsity & 65.2{\scriptsize$\pm$1.3} & 38.3{\scriptsize$\pm$0.1} & 47.3{\scriptsize$\pm$0.5} & 34.2{\scriptsize$\pm$0.8} & 19.2{\scriptsize$\pm$1.0} & 24.2{\scriptsize$\pm$0.9} & 14.9{\scriptsize$\pm$9.4} & 9.3{\scriptsize$\pm$6.7} & 11.2{\scriptsize$\pm$7.6} \\
\bottomrule
\end{tabular}
\label{tab:train_strategy}
\end{table*}

\subsection{Results}
\Cref{tab:train_strategy} reports performance over varying recovery strategies evaluated using 500 samples from each dataset.  In the top section we find that \methodabbr is largely insensitive to how it is recovered, with simple SFT being within 1.5 points of the best strategy at every budget. Training only the last three LLM blocks is the exception, but is likely due to the fact that compression error is spread across every block of both towers.  Thus, the last few blocks cannot correct it.

In \Cref{tab:train_strategy} (middle) PORTA~\cite{MinSeok2026PORTA} behaves differently, with swings in overall performance up to 29\% (compared to \methodabbr's largest overall swing of 13\%). Even SFT itself is up to 14\% off of the best method.  Further, the best recovery method changes across pruning rates, \eg, LoRA is the best at 30\%, but second worst at 10\% and 50\%.  Variance due to random seed is also generally higher than \methodabbr, with the highest being LoRA at 50\% (std of 8.1, whereas \methodabbr's highest is 2.8).  Thus, selecting the correct strategy for the pruning rate makes a significant difference for weight pruning methods like PORTA, whereas \methodabbr provides more stable and often better results.

Finally, running each source method end to end (\Cref{tab:train_strategy} (bottom)) is worse than recovering a \methodabbr or PORTA model at every budget.  Thus, we can conclude that recovery benefits may vary, with some methods being more sensitive than others (ours being less so).   However, we argue our comparisons are fair to prior work as the combination of a particular method's pruning and recovery combination does not outperform us.

\section{Setting $\gamma$ for \balancedabbr's \cref{eq:balanced}}
\label{sec:balanced_study}

\begin{figure*}[t]
\centering
\includegraphics[width=\textwidth]{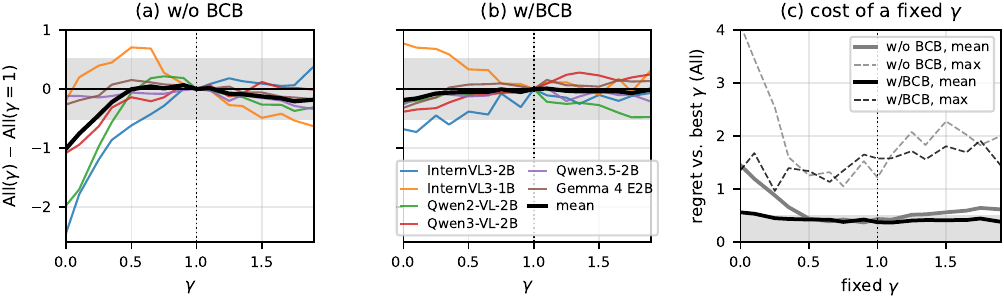}
\vspace{-8mm}
\caption{Effect of the member-balance exponent $\gamma$ (member weights $\alpha_k=\|\tilde W_k\|_F^{-\gamma}$) on six architectures, factorizing every block linear of both towers without recovery. \textbf{(a, b)} All accuracy relative to $\gamma{=}1$, averaged over the 10\%, 30\%, and 50\% \flops budgets, without and with \downstreamabbr, respectively. $\gamma{=}0$ is the unweighted joint fit. \textbf{(c)} Cost of using one fixed $\gamma$ everywhere, where the gap to the best $\gamma$ of the sweep is averaged (solid) and maximized (dashed) over the 18 [model, budget] pairs. Gray bands show the mean standard deviation.}
\label{fig:gamma}
\end{figure*}

\cref{fig:gamma} sweeps the member-balance exponent $\gamma$ on six architectures at three \flops budgets. Without \downstreamabbr (\cref{fig:gamma}a), balancing the members matters, \eg, the unweighted fit ($\gamma{=}0$) reports a 1\% drop on average compared to equal balancing $\gamma{=}1$ and up to 3.5\% in the worse case (Qwen2-VL-2B at 30\%). However, any benefit saturates quickly, with every $\gamma$ between 0.5 and 1.1 is within model variance of $\gamma{=}1$. With \downstreamabbr (\cref{fig:gamma}b), differences between $\gamma$ shrink, suggesting some redundancy with \downstreamabbr's benefits. 

\cref{fig:gamma}c measures what is lost by not tuning $\gamma$. Using $\gamma{=}1$ everywhere costs 0.5\% on average relative to the best $\gamma$ of the sweep with or without \downstreamabbr. That said, no fixed $\gamma$ performs clearly better than others in most cases. The one consistent exception is the smallest model, InternVL3-1B, which prefers weaker balancing at high compression.  For example, at 50\% \flops, $\gamma{=}0.5$ improves on $\gamma{=}1$ by just over 1\% without \downstreamabbr.  Thus, in many cases you could set $\gamma=1$ and obtain most of the gain, but maximizing performance requires setting $\gamma$ per model and \flops budget.  However, the small effect from varying $\gamma$ shows tuning only requires testing a few reasonable values (\eg, 0.5, 1, 1.5).

\section{Latency and Memory Cost Across Architectures}
\label{sec:latency}

This section describes how we measure the latency and memory costs in \cref{tab:latency}, and reports the same costs for the criteria-only results across all six architectures at the 30\% and 50\% \flops budgets (\cref{tab:latency_arch}).
\smallskip

\noindent\textbf{Measurement setup.}
All models run on a single NVIDIA L40S GPU in bfloat16 with a batch size of 1.
We evaluate latency and memory on each model on the first 500 test samples of each dataset (4{,}500 in total), using the same preprocessing and prompts as our task evaluation. However, reported task performance uses the full test set from the main results.  Each request generates exactly 16 tokens so answer length does not affect latency.
We report the mean end-to-end time per sample, covering the vision encoder, prefill and decoding, but excluding preprocessing, compilation and warm-up.
Repeated runs of the same model differ by at most 4\%.
Memory is the peak GPU memory (weights, key/value cache and activations) needed to answer the largest test sample, which is the same in every run.
\smallskip

\noindent\textbf{Deployment.}
For a fair comparison, every method, including the dense model, is deployed in the same optimized form.
Linear layers that share an input (the query/key/value and the gate/up projections) run as a single matrix multiplication, and shared-basis methods (Cat and \methodabbr) compute each shared basis $z=Bx$ once for all of its twin members (see \cref{fig:method}).
\overparamabbr factors are folded back to their dense size before deployment (described in \cref{sec:opgroups}).
Further, any factorized layer we measure slower than its dense form is also folded back into a single matrix.  A layer's widths and ranks that are not a multiple of 8 are zero-padded, which does not affect the model's outputs beyond bfloat16 rounding.
Skip connections (Cat w/ Skip and \methodabbr \& Skip) are folded into the input factor, which gives the same outputs and we found to be faster than applying them as a separate addition.
Finally, all models use a preallocated key/value cache, are compiled with \texttt{torch.compile}, and replay each decoding step as a CUDA graph.
The only exception is the prefill of Qwen3.5-2B, whose linear-attention kernels do not compile, so it runs uncompiled for every Qwen3.5-2B model, including the dense one.

\begin{table*}[t]
\centering
\small
\setlength{\tabcolsep}{3pt}
\caption{Latency and memory costs on an NVIDIA L40S GPU for the criteria-only setting of \cref{tab:architectures} (\ie,   InternVL3-2B~\cite{zhu2025internvl3} backbone using only calibration data).  We find \methodabbr reports higher performance with lower latencies and similar memory in most settings across architectures and \flops budgets. See \cref{sec:latency} for additional discussion.}
\vspace{-2mm}
\begin{tabular}{lcccccccc}
\toprule
\flops Budget & \multicolumn{4}{c}{30\%} & \multicolumn{4}{c}{50\%} \\
\cmidrule(lr){2-5} \cmidrule(lr){6-9}
 &  & Memory & Latency & Speedup &  & Memory & Latency & Speedup \\
Method & All & (GB) & (ms) & vs.\ Dense & All & (GB) & (ms) & vs.\ Dense \\
\midrule
\multicolumn{9}{l}{\emph{InternVL3-2B~\cite{zhu2025internvl3}} (dense: 58.7 All, 4.34\,GB, 135.2\,ms)} \\
SVD-LLM~\cite{wang2025svdllm} & 24.3{\scriptsize$\pm$0.5} & 3.50 & 121.3 & 1.11$\times$ & 14.9{\scriptsize$\pm$2.5} & 2.89 & 103.4 & 1.31$\times$ \\
Basis Sharing~\cite{wang2025basissharing} & 22.9{\scriptsize$\pm$0.4} & 3.42 & 123.2 & 1.10$\times$ & 13.8{\scriptsize$\pm$5.7} & 2.84 & 103.8 & 1.30$\times$ \\
Cat~\cite{lu2026skipcat} & 28.1{\scriptsize$\pm$1.1} & 3.48 & 121.0 & 1.12$\times$ & 20.3{\scriptsize$\pm$0.4} & 2.86 & 102.3 & 1.32$\times$ \\
\methodabbr (ours) & \textbf{32.2}{\scriptsize$\pm$0.9} & 3.48 & 120.9 & 1.12$\times$ & \textbf{22.2}{\scriptsize$\pm$0.5} & 2.86 & 100.5 & 1.35$\times$ \\
\midrule
\multicolumn{9}{l}{\emph{InternVL3-1B~\cite{zhu2025internvl3}} (dense: 55.0 All, 2.13\,GB, 70.3\,ms)} \\
SVD-LLM~\cite{wang2025svdllm} & 18.5{\scriptsize$\pm$0.7} & 1.84 & 69.2 & 1.02$\times$ & 3.6{\scriptsize$\pm$1.9} & 1.59 & 59.6 & 1.18$\times$ \\
Basis Sharing~\cite{wang2025basissharing} & 18.5{\scriptsize$\pm$0.3} & 1.82 & 70.3 & 1.00$\times$ & 4.8{\scriptsize$\pm$0.1} & 1.58 & 60.9 & 1.15$\times$ \\
Cat~\cite{lu2026skipcat} & 21.1{\scriptsize$\pm$0.7} & 1.83 & 67.9 & 1.04$\times$ & 18.2{\scriptsize$\pm$0.3} & 1.57 & 62.4 & 1.13$\times$ \\
\methodabbr (ours) & \textbf{23.1}{\scriptsize$\pm$0.7} & 1.83 & 68.7 & 1.02$\times$ & \textbf{18.5}{\scriptsize$\pm$0.1} & 1.57 & 63.4 & 1.11$\times$ \\
\midrule
\multicolumn{9}{l}{\emph{Qwen2-VL-2B~\cite{wang2024qwen2vl}} (dense: 59.1 All, 4.28\,GB, 133.5\,ms)} \\
SVD-LLM~\cite{wang2025svdllm} & 25.5{\scriptsize$\pm$0.1} & 3.24 & 135.6 & 0.98$\times$ & 18.1{\scriptsize$\pm$0.8} & 2.49 & 123.9 & 1.08$\times$ \\
Basis Sharing~\cite{wang2025basissharing} & 24.3{\scriptsize$\pm$0.2} & 3.17 & 133.1 & 1.00$\times$ & 17.2{\scriptsize$\pm$1.2} & 2.44 & 122.4 & 1.09$\times$ \\
Cat~\cite{lu2026skipcat} & 27.0{\scriptsize$\pm$0.4} & 3.23 & 133.1 & 1.00$\times$ & 21.1{\scriptsize$\pm$0.2} & 2.47 & 119.1 & 1.12$\times$ \\
\methodabbr (ours) & \textbf{31.8}{\scriptsize$\pm$0.9} & 3.23 & 131.4 & 1.02$\times$ & \textbf{21.8}{\scriptsize$\pm$0.1} & 2.47 & 121.8 & 1.10$\times$ \\
\midrule
\multicolumn{9}{l}{\emph{Qwen3-VL-2B~\cite{bai2025qwen3vl}} (dense: 58.1 All, 4.24\,GB, 141.8\,ms)} \\
SVD-LLM~\cite{wang2025svdllm} & \textbf{25.8}{\scriptsize$\pm$0.2} & 3.31 & 139.6 & 1.02$\times$ & 19.9{\scriptsize$\pm$0.6} & 2.65 & 125.8 & 1.13$\times$ \\
Basis Sharing~\cite{wang2025basissharing} & 24.5{\scriptsize$\pm$0.6} & 3.20 & 140.1 & 1.01$\times$ & 19.4{\scriptsize$\pm$0.4} & 2.58 & 128.8 & 1.10$\times$ \\
Cat~\cite{lu2026skipcat} & 24.8{\scriptsize$\pm$0.3} & 3.31 & 133.5 & 1.06$\times$ & 20.3{\scriptsize$\pm$0.2} & 2.66 & 124.1 & 1.14$\times$ \\
\methodabbr (ours) & 25.6{\scriptsize$\pm$0.4} & 3.31 & 135.2 & 1.05$\times$ & \textbf{21.0}{\scriptsize$\pm$0.3} & 2.66 & 123.2 & 1.15$\times$ \\
\midrule
\multicolumn{9}{l}{\emph{Qwen3.5-2B~\cite{qwen2026qwen35}} (dense: 57.1 All, 4.27\,GB, 162.7\,ms)} \\
SVD-LLM~\cite{wang2025svdllm} & 24.7{\scriptsize$\pm$0.2} & 3.36 & 157.9 & 1.03$\times$ & 18.4{\scriptsize$\pm$0.2} & 2.71 & 145.1 & 1.12$\times$ \\
Basis Sharing~\cite{wang2025basissharing} & 24.6{\scriptsize$\pm$0.4} & 3.27 & 156.4 & 1.04$\times$ & 18.6{\scriptsize$\pm$0.7} & 2.65 & 146.6 & 1.11$\times$ \\
Cat~\cite{lu2026skipcat} & \textbf{25.0}{\scriptsize$\pm$0.3} & 3.35 & 154.5 & 1.05$\times$ & 19.3{\scriptsize$\pm$0.0} & 2.71 & 142.9 & 1.14$\times$ \\
\methodabbr (ours) & 24.8{\scriptsize$\pm$0.4} & 3.35 & 153.2 & 1.06$\times$ & \textbf{19.8}{\scriptsize$\pm$0.1} & 2.71 & 142.9 & 1.14$\times$ \\
\midrule
\multicolumn{9}{l}{\emph{Gemma 4 E2B~\cite{gemmateam2026gemma4}} (dense: 44.6 All, 9.70\,GB, 157.4\,ms)} \\
SVD-LLM~\cite{wang2025svdllm} & 19.6{\scriptsize$\pm$0.3} & 8.63 & 142.7 & 1.10$\times$ & 14.0{\scriptsize$\pm$0.1} & 7.91 & 129.2 & 1.22$\times$ \\
AdaSVD~\cite{Li_2026_CVPR_Adasvd} & 21.7{\scriptsize$\pm$0.1} & 8.70 & 148.0 & 1.06$\times$ & 14.8{\scriptsize$\pm$0.3} & 7.92 & 134.1 & 1.17$\times$ \\
Basis Sharing~\cite{wang2025basissharing} & 18.8{\scriptsize$\pm$0.4} & 8.55 & 148.7 & 1.06$\times$ & 13.7{\scriptsize$\pm$0.0} & 7.85 & 131.9 & 1.19$\times$ \\
Cat~\cite{lu2026skipcat} & 23.0{\scriptsize$\pm$0.4} & 8.62 & 142.0 & 1.11$\times$ & \textbf{16.2}{\scriptsize$\pm$0.1} & 7.89 & 128.7 & 1.22$\times$ \\
\methodabbr (ours) & \textbf{23.6}{\scriptsize$\pm$0.3} & 8.62 & 143.1 & 1.10$\times$ & 15.5{\scriptsize$\pm$0.2} & 7.89 & 128.1 & 1.23$\times$ \\
\bottomrule
\end{tabular}
\vspace{-2mm}
\label{tab:latency_arch}
\end{table*}

\subsection{Results}

\cref{tab:latency_arch} reports the latency and memory costs across architectures to augment the method-focused evaluation in \cref{tab:latency} of the main paper.  Generally speaking, our method gets the highest performance with the lowest latencies between all the \taskabbr methods in our comparisons.  Basis Sharing~\cite{wang2025basissharing} sometimes requires a little less memory, but comes at significant loss in task accuracy that can be more than made up by reducing the \flops budget.  For example, both InternVL3 models have 30\% \flops Basis Sharing accuracies that match \methodabbr at 50\%, but where ours has much lower latencies at that budget. Further, \cref{tab:latency_arch} shows the factorized models at 50\% are 1.08--1.35$\times$ faster than the dense model and use 18--43\% less memory, demonstrating that the computational gains can be reached in practice.  That said, they do not achieve the 50\% speedup their \flops budget would suggest, highlighting a need for better support for these types of methods to obtain their full benefits.


\section{Per-dataset Results}
\label{sec:perdataset}

\cref{tab:criteria_perdataset,tab:strategy_perdataset,tab:recovery_perdataset,tab:architectures_perdataset,tab:balanced_perdataset,tab:plusprune_perdataset,tab:plusquant_perdataset,tab:plustok_perdataset} break the summary tables from the main paper down by dataset.  We provide a discussion below:
\smallskip

\begin{table*}[t]
\centering
\scriptsize
\setlength{\tabcolsep}{2pt}
\caption{Per-dataset results of the criteria-only evaluation in \cref{tab:criteria}.  See \cref{sec:perdataset} for discussion.}
\resizebox{\textwidth}{!}{%
\begin{tabular}{llccc|cccccc|c}
\toprule
\flops & & \multicolumn{3}{c|}{Seen} & \multicolumn{6}{c|}{Unseen} & \\
Budget & Method & CLEVRER & GQA & NExT-QA & Perc.Test & TextVQA & VizWiz & VQAv2 & VideoMME & LVB & All \\
\midrule
\multirow{15}{*}{30\%} & FLAP~\cite{an2024flap} & 3.2{\scriptsize$\pm$0.2} & 0.8{\scriptsize$\pm$0.7} & 15.5{\scriptsize$\pm$2.7} & 27.8{\scriptsize$\pm$3.0} & 0.1{\scriptsize$\pm$0.0} & 0.0{\scriptsize$\pm$0.0} & 1.3{\scriptsize$\pm$0.9} & 22.1{\scriptsize$\pm$0.8} & 19.8{\scriptsize$\pm$0.1} & 10.1{\scriptsize$\pm$0.6} \\
 & Wanda~\cite{sun2024wanda} & 0.4{\scriptsize$\pm$0.1} & 0.1{\scriptsize$\pm$0.0} & 2.1{\scriptsize$\pm$1.1} & 2.2{\scriptsize$\pm$1.0} & 0.2{\scriptsize$\pm$0.1} & 0.0{\scriptsize$\pm$0.1} & 0.1{\scriptsize$\pm$0.0} & 1.8{\scriptsize$\pm$0.9} & 2.9{\scriptsize$\pm$1.0} & 1.1{\scriptsize$\pm$0.4} \\
 & Putri~\cite{coello2026putri} & 1.6{\scriptsize$\pm$1.6} & 1.2{\scriptsize$\pm$2.0} & 4.4{\scriptsize$\pm$4.7} & 3.2{\scriptsize$\pm$4.0} & 0.3{\scriptsize$\pm$0.3} & 0.0{\scriptsize$\pm$0.0} & 2.2{\scriptsize$\pm$3.7} & 6.7{\scriptsize$\pm$7.8} & 7.1{\scriptsize$\pm$5.6} & 3.0{\scriptsize$\pm$2.3} \\
 & Wanda~\cite{sun2024wanda}+Putri~\cite{coello2026putri} & 27.7{\scriptsize$\pm$1.3} & 21.8{\scriptsize$\pm$0.4} & 26.2{\scriptsize$\pm$1.3} & 41.3{\scriptsize$\pm$0.6} & 7.3{\scriptsize$\pm$0.5} & 7.5{\scriptsize$\pm$0.3} & 32.8{\scriptsize$\pm$0.7} & 27.3{\scriptsize$\pm$0.4} & 27.5{\scriptsize$\pm$0.4} & 24.4{\scriptsize$\pm$0.5} \\
 & AMP~\cite{mugnaini2025amp} & 0.6{\scriptsize$\pm$0.2} & 0.0{\scriptsize$\pm$0.0} & 1.9{\scriptsize$\pm$0.5} & 3.4{\scriptsize$\pm$0.2} & 0.2{\scriptsize$\pm$0.1} & 0.0{\scriptsize$\pm$0.0} & 0.1{\scriptsize$\pm$0.0} & 0.8{\scriptsize$\pm$0.3} & 1.9{\scriptsize$\pm$0.8} & 1.0{\scriptsize$\pm$0.1} \\
\bottomrule
\end{tabular}}
\label{tab:criteria_perdataset}
\end{table*}

\begin{table*}[t]
\centering
\scriptsize
\setlength{\tabcolsep}{2pt}
\caption{Per-dataset results comparing method-specific budget allocators from \cref{tab:strategy}. See \cref{sec:perdataset} for discussion.}
\resizebox{\textwidth}{!}{%
\begin{tabular}{llccc|cccccc|c}
\toprule
 \flops & & \multicolumn{3}{c|}{Seen} & \multicolumn{6}{c|}{Unseen} & \\
Budget & Method & CLEVRER & GQA & NExT-QA & Perc.Test & TextVQA & VizWiz & VQAv2 & VideoMME & LVB & All \\
\midrule
\multirow{8}{*}{10\%} & LLM-Pruner~\cite{ma2023llmpruner} & 20.7{\scriptsize$\pm$0.5} & 40.5{\scriptsize$\pm$0.3} & 44.9{\scriptsize$\pm$0.6} & 41.5{\scriptsize$\pm$0.1} & 36.1{\scriptsize$\pm$0.3} & 43.6{\scriptsize$\pm$1.0} & 58.8{\scriptsize$\pm$0.3} & 36.0{\scriptsize$\pm$0.5} & 34.7{\scriptsize$\pm$0.3} & 39.6{\scriptsize$\pm$0.1} \\
 & SlimLLM~\cite{guo2025slimllm} & 26.5{\scriptsize$\pm$0.6} & 48.5{\scriptsize$\pm$0.2} & 63.0{\scriptsize$\pm$0.3} & 44.0{\scriptsize$\pm$0.2} & 61.1{\scriptsize$\pm$0.3} & \textbf{50.8}{\scriptsize$\pm$1.5} & 69.4{\scriptsize$\pm$0.1} & \textbf{41.3}{\scriptsize$\pm$0.3} & \textbf{44.9}{\scriptsize$\pm$0.5} & 49.9{\scriptsize$\pm$0.2} \\
 & PruneLLM~\cite{yang2025prunellm} & 29.5{\scriptsize$\pm$0.3} & 43.3{\scriptsize$\pm$0.2} & 58.3{\scriptsize$\pm$0.5} & 35.3{\scriptsize$\pm$0.7} & \textbf{64.2}{\scriptsize$\pm$0.4} & 43.7{\scriptsize$\pm$0.4} & 66.3{\scriptsize$\pm$0.3} & 36.1{\scriptsize$\pm$0.6} & 40.5{\scriptsize$\pm$0.4} & 46.4{\scriptsize$\pm$0.2} \\
 & Putri~\cite{coello2026putri} & 15.9{\scriptsize$\pm$6.9} & 29.8{\scriptsize$\pm$3.0} & 35.6{\scriptsize$\pm$4.5} & 39.1{\scriptsize$\pm$2.5} & 31.2{\scriptsize$\pm$1.4} & 10.8{\scriptsize$\pm$3.7} & 44.9{\scriptsize$\pm$3.2} & 28.2{\scriptsize$\pm$1.9} & 26.4{\scriptsize$\pm$1.9} & 29.1{\scriptsize$\pm$1.7} \\
 & TRSP~\cite{feng2026trsp} & 6.7{\scriptsize$\pm$1.9} & 32.8{\scriptsize$\pm$2.2} & 37.6{\scriptsize$\pm$4.6} & 42.5{\scriptsize$\pm$1.0} & 45.0{\scriptsize$\pm$4.7} & 30.7{\scriptsize$\pm$2.3} & 43.5{\scriptsize$\pm$2.9} & 34.2{\scriptsize$\pm$1.0} & 26.1{\scriptsize$\pm$2.1} & 33.2{\scriptsize$\pm$2.2} \\
\bottomrule
\end{tabular}}
\label{tab:strategy_perdataset}
\end{table*}

\begin{table*}[t]
\centering
\scriptsize
\caption{Per-dataset results after performing recovery using the Seen datasets.  \textbf{(a)} runs each method's own compression and recovery as described in its paper (not shown in \cref{tab:recovery}), \textbf{(b)} reports the per-dataset SFT results for \cref{tab:recovery}. See \cref{sec:perdataset} for discussion.}
\setlength{\tabcolsep}{2pt}
\resizebox{\textwidth}{!}{%
\begin{tabular}{llccc|cccccc|c}
\toprule
\flops & & \multicolumn{3}{c|}{Seen} & \multicolumn{6}{c|}{Unseen} & \\
Budget & Method & CLEVRER & GQA & NExT-QA & Perc.Test & TextVQA & VizWiz & VQAv2 & VideoMME & LVB & All \\
\midrule
\multicolumn{12}{l}{\textbf{(a) \emph{Source method end to end}}} \\
\midrule
\multirow{9}{*}{30\%} & AMP~\cite{mugnaini2025amp} & \textbf{69.9}{\scriptsize$\pm$0.2} & 52.5{\scriptsize$\pm$6.2} & \textbf{62.7}{\scriptsize$\pm$0.1} & 39.8{\scriptsize$\pm$5.0} & 9.4{\scriptsize$\pm$0.3} & 11.6{\scriptsize$\pm$1.5} & 51.9{\scriptsize$\pm$5.3} & 31.5{\scriptsize$\pm$0.7} & \textbf{34.2}{\scriptsize$\pm$0.9} & 40.4{\scriptsize$\pm$1.6} \\
 & LLM-Pruner~\cite{ma2023llmpruner} & 50.0{\scriptsize$\pm$1.7} & 55.3{\scriptsize$\pm$1.1} & 42.9{\scriptsize$\pm$4.4} & 32.3{\scriptsize$\pm$3.2} & 20.0{\scriptsize$\pm$2.3} & 17.9{\scriptsize$\pm$0.3} & 58.5{\scriptsize$\pm$1.8} & 29.8{\scriptsize$\pm$0.9} & 29.7{\scriptsize$\pm$1.5} & 37.4{\scriptsize$\pm$1.1} \\
 & SlimLLM~\cite{guo2025slimllm} & 57.0{\scriptsize$\pm$0.3} & \textbf{58.0}{\scriptsize$\pm$0.3} & 56.7{\scriptsize$\pm$0.5} & 37.7{\scriptsize$\pm$3.0} & \textbf{25.0}{\scriptsize$\pm$0.9} & \textbf{29.6}{\scriptsize$\pm$10.8} & \textbf{61.2}{\scriptsize$\pm$0.2} & 31.4{\scriptsize$\pm$1.0} & 33.3{\scriptsize$\pm$0.8} & \textbf{43.3}{\scriptsize$\pm$0.7} \\
 & SVD-LLM~\cite{wang2025svdllm} & 48.7{\scriptsize$\pm$4.5} & 53.0{\scriptsize$\pm$1.4} & 51.9{\scriptsize$\pm$0.8} & 38.1{\scriptsize$\pm$2.0} & 13.2{\scriptsize$\pm$1.6} & 14.9{\scriptsize$\pm$0.8} & 53.9{\scriptsize$\pm$1.3} & 31.0{\scriptsize$\pm$1.3} & 33.4{\scriptsize$\pm$1.6} & 37.6{\scriptsize$\pm$0.9} \\
 & Minitron~\cite{muralidharan2024minitron} & 31.8{\scriptsize$\pm$1.1} & 50.2{\scriptsize$\pm$1.0} & 53.0{\scriptsize$\pm$0.1} & \textbf{46.1}{\scriptsize$\pm$0.9} & 22.8{\scriptsize$\pm$0.4} & 25.2{\scriptsize$\pm$7.5} & 56.0{\scriptsize$\pm$1.9} & 30.5{\scriptsize$\pm$0.5} & 32.7{\scriptsize$\pm$0.5} & 38.7{\scriptsize$\pm$1.2} \\
\bottomrule
\end{tabular}}
\label{tab:recovery_perdataset}
\end{table*}

\begin{table*}[t]
\centering
\scriptsize
\setlength{\tabcolsep}{2pt}
\caption{Per-dataset results for the criteria-only evaluation across architectures in \cref{tab:architectures}. See \cref{sec:perdataset} for discussion.}
\resizebox{0.75\textwidth}{!}{%
\begin{tabular}{llccc|cccccc|c}
\toprule
 & & \multicolumn{3}{c|}{Seen} & \multicolumn{6}{c|}{Unseen} & \\
Budget & Method & CLEVRER & GQA & NExT-QA & Perc.Test & TextVQA & VizWiz & VQAv2 & VideoMME & LVB & All \\
\midrule
\multirow{26}{*}{10\%} & \multicolumn{11}{l}{\emph{InternVL3-1B}} \\
 & SVD-LLM~\cite{wang2025svdllm} & 26.4{\scriptsize$\pm$0.4} & 34.7{\scriptsize$\pm$1.4} & 38.3{\scriptsize$\pm$0.5} & 40.0{\scriptsize$\pm$0.8} & 19.3{\scriptsize$\pm$0.3} & 24.4{\scriptsize$\pm$7.4} & 45.4{\scriptsize$\pm$2.0} & 29.2{\scriptsize$\pm$0.9} & 29.2{\scriptsize$\pm$1.3} & 31.9{\scriptsize$\pm$0.5} \\
 & Basis Sharing~\cite{wang2025basissharing} & 27.4{\scriptsize$\pm$1.1} & 32.8{\scriptsize$\pm$0.6} & 34.1{\scriptsize$\pm$0.3} & 40.4{\scriptsize$\pm$0.6} & 14.8{\scriptsize$\pm$1.4} & 11.1{\scriptsize$\pm$0.4} & 44.7{\scriptsize$\pm$0.6} & 29.2{\scriptsize$\pm$0.3} & 26.6{\scriptsize$\pm$0.8} & 29.0{\scriptsize$\pm$0.3} \\
 & Cat~\cite{lu2026skipcat} & \textbf{32.0}{\scriptsize$\pm$0.7} & 40.2{\scriptsize$\pm$0.6} & 46.8{\scriptsize$\pm$1.3} & 42.1{\scriptsize$\pm$0.4} & \textbf{26.8}{\scriptsize$\pm$0.5} & 30.7{\scriptsize$\pm$12.7} & 53.2{\scriptsize$\pm$0.4} & 31.7{\scriptsize$\pm$0.3} & 31.2{\scriptsize$\pm$0.6} & 37.2{\scriptsize$\pm$1.6} \\
 & \methodabbr (ours) & \textbf{32.0}{\scriptsize$\pm$0.5} & \textbf{40.8}{\scriptsize$\pm$0.4} & \textbf{47.7}{\scriptsize$\pm$0.7} & \textbf{42.4}{\scriptsize$\pm$0.6} & 26.5{\scriptsize$\pm$0.4} & \textbf{36.5}{\scriptsize$\pm$6.1} & \textbf{53.5}{\scriptsize$\pm$0.3} & \textbf{32.1}{\scriptsize$\pm$0.6} & \textbf{32.7}{\scriptsize$\pm$0.5} & \textbf{38.2}{\scriptsize$\pm$0.5} \\
 & \multicolumn{11}{l}{\emph{Qwen2-VL-2B}} \\
\bottomrule
\end{tabular}}
\label{tab:architectures_perdataset}
\end{table*}

\noindent\textbf{Datasets differ greatly in their sensitivity to compression.} Open-ended image question answering (TextVQA, VizWiz, VQAv2, GQA) degrades the fastest across tables, with TextVQA being the most sensitive, which requires reading fine-grained text in the image. In \cref{tab:criteria_perdataset}, TextVQA falls up to 8\% for the best methods, whereas Perception Test only loses 2.5\% for \methodabbr after 20\% additional compression. Although multiple-choice video benchmarks may appear to degrade more slowly, this is due, in part, to the fact that they also have a chance floor (33.3\% for Perception Test and 20\% for NExT-QA). Without recovery, many methods sit near that floor at the 50\% budget (see \cref{tab:architectures_perdataset}). Thus, many of the differences without recovery in All at the highest \flops rates come mostly from the open-ended datasets.
\smallskip

\noindent\textbf{Budget allocation.} We found that methods that keep the whole vision tower intact helps at low compression.  For example, at 10\% SlimLLM is the best method on VizWiz, VideoMME, and LongVideoBench, even though it is 3\% lower on All (\cref{tab:strategy_perdataset}). The same choice breaks down at 30\%, where removing 30\% of the model's \flops from the LLM alone collapses most LLM-only methods.
\smallskip

\noindent\textbf{Recovery.} \cref{tab:recovery_perdataset} shows most of the Seen gain from recovery comes from CLEVRER.  This is due, in part, to the poor performance from the base model itself (\cref{tab:base_perdataset} reports a 45\% gain from SFT on the dense model).  In fact, the dense model loses accuracy on some datasets after SFT (\eg, TextVQA, VQAv2, VideoMME). In contrast, at 50\% \flops recovery significantly helps Unseen VQAv2 accuracy, whose questions resemble GQA's from the recovery mix, while TextVQA gets a much smaller gain. This suggests that recovery restores skills Seen datasets exercise but not skills they lack, such as reading text. The source methods in \cref{tab:recovery_perdataset}(a) show the same trade-off.
\smallskip

\noindent\textbf{Other architectures.} \methodabbr has the best All score in 12 of the 15 model-budget pairs in \cref{tab:architectures_perdataset}, and its largest gains come from datasets where the other \taskabbr methods collapse. For example, on Qwen2-VL-2B at 30\% it improves All by 5\% over Cat, mainly on NExT-QA (9.5\% gain) and VizWiz (10\% gain). The three pairs where \methodabbr is second (Qwen3-VL-2B and Qwen3.5-2B at 30\%, Gemma 4 E2B at 50\%) are also the ones where every method is close.  \Ie, the spread in All is at most 2.5 points and several multiple-choice datasets are near chance for every method.

\begin{table*}[t]
\centering
\scriptsize
\setlength{\tabcolsep}{2pt}
\caption{Per-dataset results comparing \balancedabbr with BKV~\cite{meng2025transmla} from \cref{tab:balanced}. See \cref{sec:perdataset} for discussion.}
\resizebox{\textwidth}{!}{%
\begin{tabular}{llccc|cccccc|c}
\toprule
  \flops & & \multicolumn{3}{c|}{Seen} & \multicolumn{6}{c|}{Unseen} & \\
Budget & Method & CLEVRER & GQA & NExT-QA & Perc.Test & TextVQA & VizWiz & VQAv2 & VideoMME & LVB & All \\
\midrule
\multirow{2}{*}{10\%} & BKV~\cite{meng2025transmla} & \textbf{32.2}{\scriptsize$\pm$0.2} & 40.7{\scriptsize$\pm$0.2} & 47.5{\scriptsize$\pm$1.4} & \textbf{43.4}{\scriptsize$\pm$0.5} & 25.8{\scriptsize$\pm$0.3} & 31.5{\scriptsize$\pm$9.1} & 53.3{\scriptsize$\pm$0.5} & 31.6{\scriptsize$\pm$0.2} & 31.6{\scriptsize$\pm$1.2} & 37.5{\scriptsize$\pm$0.7} \\
 & \balancedabbr (ours) & 32.0{\scriptsize$\pm$0.5} & \textbf{40.8}{\scriptsize$\pm$0.4} & \textbf{47.7}{\scriptsize$\pm$0.7} & 42.4{\scriptsize$\pm$0.6} & \textbf{26.5}{\scriptsize$\pm$0.4} & \textbf{36.5}{\scriptsize$\pm$6.1} & \textbf{53.5}{\scriptsize$\pm$0.3} & \textbf{32.1}{\scriptsize$\pm$0.6} & \textbf{32.7}{\scriptsize$\pm$0.5} & \textbf{38.2}{\scriptsize$\pm$0.5} \\
\midrule
\multirow{2}{*}{30\%} & BKV~\cite{meng2025transmla} & \textbf{28.6}{\scriptsize$\pm$0.2} & \textbf{24.4}{\scriptsize$\pm$0.9} & \textbf{26.8}{\scriptsize$\pm$0.9} & \textbf{37.7}{\scriptsize$\pm$0.6} & 5.4{\scriptsize$\pm$0.2} & 4.2{\scriptsize$\pm$0.4} & \textbf{34.0}{\scriptsize$\pm$0.7} & \textbf{26.4}{\scriptsize$\pm$0.3} & 24.4{\scriptsize$\pm$0.8} & \textbf{23.5}{\scriptsize$\pm$0.2} \\
 & \balancedabbr (ours) & 27.3{\scriptsize$\pm$0.6} & 22.4{\scriptsize$\pm$2.6} & 26.4{\scriptsize$\pm$0.1} & 36.7{\scriptsize$\pm$0.1} & \textbf{5.8}{\scriptsize$\pm$0.3} & \textbf{4.7}{\scriptsize$\pm$0.8} & 33.1{\scriptsize$\pm$1.7} & 25.9{\scriptsize$\pm$0.7} & \textbf{25.5}{\scriptsize$\pm$1.2} & 23.1{\scriptsize$\pm$0.7} \\
\midrule
\multirow{2}{*}{50\%} & BKV~\cite{meng2025transmla} & 11.7{\scriptsize$\pm$0.6} & 11.6{\scriptsize$\pm$1.2} & 20.1{\scriptsize$\pm$0.2} & \textbf{33.4}{\scriptsize$\pm$0.7} & \textbf{4.5}{\scriptsize$\pm$0.3} & 4.0{\scriptsize$\pm$0.6} & 24.2{\scriptsize$\pm$1.7} & 25.6{\scriptsize$\pm$0.4} & 21.4{\scriptsize$\pm$0.1} & 17.4{\scriptsize$\pm$0.3} \\
 & \balancedabbr (ours) & \textbf{14.0}{\scriptsize$\pm$0.1} & \textbf{13.2}{\scriptsize$\pm$0.5} & \textbf{20.4}{\scriptsize$\pm$0.6} & 31.4{\scriptsize$\pm$0.5} & 4.2{\scriptsize$\pm$0.7} & \textbf{6.9}{\scriptsize$\pm$0.2} & \textbf{27.8}{\scriptsize$\pm$0.2} & \textbf{26.4}{\scriptsize$\pm$0.2} & \textbf{22.1}{\scriptsize$\pm$0.6} & \textbf{18.5}{\scriptsize$\pm$0.1} \\
\bottomrule
\end{tabular}}
\label{tab:balanced_perdataset}
\end{table*}

\begin{table*}[t]
\centering
\scriptsize
\setlength{\tabcolsep}{2pt}
\caption{Per-dataset results on joint \taskabbr and weight pruning from \cref{tab:plusprune}. See \cref{sec:perdataset} for discussion.}
\resizebox{\textwidth}{!}{%
\begin{tabular}{llccc|cccccc|c}
\toprule
  \flops& & \multicolumn{3}{c|}{Seen} & \multicolumn{6}{c|}{Unseen} & \\
Budget & Method & CLEVRER & GQA & NExT-QA & Perc.Test & TextVQA & VizWiz & VQAv2 & VideoMME & LVB & All \\
\midrule
\multirow{4}{*}{10\%} & Cat w/Skip~\cite{lu2026skipcat} & \textbf{37.2}{\scriptsize$\pm$1.2} & 55.5{\scriptsize$\pm$0.1} & 70.4{\scriptsize$\pm$0.5} & 51.1{\scriptsize$\pm$1.0} & 68.9{\scriptsize$\pm$0.4} & 46.8{\scriptsize$\pm$3.4} & 73.8{\scriptsize$\pm$0.1} & \textbf{43.4}{\scriptsize$\pm$0.4} & 43.7{\scriptsize$\pm$0.7} & 54.5{\scriptsize$\pm$0.7} \\
 & \methodabbr (ours) & 36.1{\scriptsize$\pm$0.4} & 48.6{\scriptsize$\pm$0.4} & 61.5{\scriptsize$\pm$0.3} & 47.1{\scriptsize$\pm$0.5} & 47.7{\scriptsize$\pm$1.7} & 43.4{\scriptsize$\pm$2.3} & 64.1{\scriptsize$\pm$0.1} & 37.7{\scriptsize$\pm$0.0} & 41.6{\scriptsize$\pm$1.2} & 47.5{\scriptsize$\pm$0.3} \\
 & \methodabbr w/Skip & \textbf{37.2}{\scriptsize$\pm$0.3} & \textbf{56.2}{\scriptsize$\pm$0.4} & \textbf{71.8}{\scriptsize$\pm$0.2} & \textbf{52.1}{\scriptsize$\pm$0.5} & 70.0{\scriptsize$\pm$0.5} & 46.6{\scriptsize$\pm$1.6} & \textbf{74.0}{\scriptsize$\pm$0.3} & 43.1{\scriptsize$\pm$0.4} & 44.7{\scriptsize$\pm$0.9} & 55.1{\scriptsize$\pm$0.4} \\
 & \methodabbr w/\overparamabbr w/Skip & \textbf{37.2}{\scriptsize$\pm$0.5} & 56.1{\scriptsize$\pm$0.1} & 71.3{\scriptsize$\pm$0.4} & 50.6{\scriptsize$\pm$0.4} & \textbf{70.2}{\scriptsize$\pm$0.5} & \textbf{51.3}{\scriptsize$\pm$1.1} & 73.9{\scriptsize$\pm$0.2} & 43.0{\scriptsize$\pm$0.7} & \textbf{44.9}{\scriptsize$\pm$0.7} & \textbf{55.4}{\scriptsize$\pm$0.2} \\
\midrule
\multirow{4}{*}{30\%} & Cat w/Skip~\cite{lu2026skipcat} & 32.7{\scriptsize$\pm$1.1} & 38.2{\scriptsize$\pm$0.3} & 41.7{\scriptsize$\pm$1.5} & 41.5{\scriptsize$\pm$0.6} & 17.3{\scriptsize$\pm$0.6} & 40.7{\scriptsize$\pm$4.1} & 49.9{\scriptsize$\pm$0.4} & 31.2{\scriptsize$\pm$0.9} & 33.5{\scriptsize$\pm$0.7} & 36.3{\scriptsize$\pm$0.8} \\
 & \methodabbr (ours) & \textbf{33.1}{\scriptsize$\pm$1.1} & 32.0{\scriptsize$\pm$1.0} & 38.1{\scriptsize$\pm$0.9} & 41.3{\scriptsize$\pm$1.0} & 14.3{\scriptsize$\pm$0.7} & 29.2{\scriptsize$\pm$7.7} & 40.6{\scriptsize$\pm$0.5} & 29.5{\scriptsize$\pm$0.7} & 32.1{\scriptsize$\pm$2.1} & 32.2{\scriptsize$\pm$0.9} \\
 & \methodabbr w/Skip & 32.3{\scriptsize$\pm$0.7} & 40.2{\scriptsize$\pm$0.3} & 47.4{\scriptsize$\pm$1.2} & 43.1{\scriptsize$\pm$0.8} & 23.6{\scriptsize$\pm$1.0} & \textbf{46.7}{\scriptsize$\pm$1.6} & 52.2{\scriptsize$\pm$0.5} & 32.3{\scriptsize$\pm$0.9} & \textbf{36.3}{\scriptsize$\pm$0.2} & 39.3{\scriptsize$\pm$0.5} \\
 & \methodabbr w/\overparamabbr w/Skip & 32.9{\scriptsize$\pm$0.8} & \textbf{41.5}{\scriptsize$\pm$0.8} & \textbf{48.3}{\scriptsize$\pm$0.6} & \textbf{44.3}{\scriptsize$\pm$0.9} & \textbf{29.9}{\scriptsize$\pm$0.7} & 36.3{\scriptsize$\pm$5.4} & \textbf{56.0}{\scriptsize$\pm$0.4} & \textbf{34.7}{\scriptsize$\pm$0.5} & 33.8{\scriptsize$\pm$0.5} & \textbf{39.7}{\scriptsize$\pm$0.6} \\
\midrule
\multirow{4}{*}{50\%} & Cat w/Skip~\cite{lu2026skipcat} & 26.2{\scriptsize$\pm$0.5} & 21.7{\scriptsize$\pm$0.3} & 22.9{\scriptsize$\pm$0.8} & 36.5{\scriptsize$\pm$1.0} & 6.0{\scriptsize$\pm$0.2} & 5.7{\scriptsize$\pm$0.6} & 29.7{\scriptsize$\pm$0.5} & 26.5{\scriptsize$\pm$1.2} & 23.3{\scriptsize$\pm$0.7} & 22.0{\scriptsize$\pm$0.4} \\
 & \methodabbr (ours) & 25.4{\scriptsize$\pm$1.4} & 21.6{\scriptsize$\pm$0.1} & 22.9{\scriptsize$\pm$0.8} & 38.8{\scriptsize$\pm$0.9} & 6.2{\scriptsize$\pm$0.1} & 5.5{\scriptsize$\pm$0.1} & 29.7{\scriptsize$\pm$0.2} & 26.2{\scriptsize$\pm$0.8} & 23.4{\scriptsize$\pm$1.4} & 22.2{\scriptsize$\pm$0.5} \\
 & \methodabbr w/Skip & \textbf{27.2}{\scriptsize$\pm$1.2} & 22.9{\scriptsize$\pm$0.3} & 24.8{\scriptsize$\pm$0.6} & \textbf{39.3}{\scriptsize$\pm$0.9} & 7.2{\scriptsize$\pm$0.2} & 6.3{\scriptsize$\pm$0.2} & 31.3{\scriptsize$\pm$0.3} & 26.6{\scriptsize$\pm$0.3} & \textbf{24.0}{\scriptsize$\pm$0.8} & 23.3{\scriptsize$\pm$0.3} \\
 & \methodabbr w/\overparamabbr w/Skip & 26.2{\scriptsize$\pm$0.8} & \textbf{23.6}{\scriptsize$\pm$0.1} & \textbf{24.9}{\scriptsize$\pm$0.6} & 37.1{\scriptsize$\pm$0.3} & \textbf{7.8}{\scriptsize$\pm$0.2} & \textbf{8.5}{\scriptsize$\pm$0.6} & \textbf{34.9}{\scriptsize$\pm$0.4} & \textbf{27.6}{\scriptsize$\pm$0.7} & 23.4{\scriptsize$\pm$1.3} & \textbf{23.8}{\scriptsize$\pm$0.1} \\
\bottomrule
\end{tabular}}
\label{tab:plusprune_perdataset}
\end{table*}

\begin{table*}[t]
\centering
\scriptsize
\setlength{\tabcolsep}{2pt}
\caption{Per-dataset results on joint \taskabbr and quantization from \cref{tab:plusquant}. See \cref{sec:perdataset} for discussion.}
\resizebox{\textwidth}{!}{%
\begin{tabular}{llccc|cccccc|c}
\toprule
  \flops & & \multicolumn{3}{c|}{Seen} & \multicolumn{6}{c|}{Unseen} & \\
Budget & Method & CLEVRER & GQA & NExT-QA & Perc.Test & TextVQA & VizWiz & VQAv2 & VideoMME & LVB & All \\
\midrule
\multirow{3}{*}{50\%} & GETA~\cite{qu2025geta} & 60.9{\scriptsize$\pm$1.0} & 39.9{\scriptsize$\pm$0.2} & 45.6{\scriptsize$\pm$0.8} & 28.6{\scriptsize$\pm$3.6} & \textbf{5.7}{\scriptsize$\pm$0.1} & 8.6{\scriptsize$\pm$0.8} & 40.1{\scriptsize$\pm$0.3} & 25.6{\scriptsize$\pm$0.5} & 27.9{\scriptsize$\pm$0.8} & 31.4{\scriptsize$\pm$0.5} \\
 & UniQL~\cite{chiang2026uniql} & 74.9{\scriptsize$\pm$1.0} & 44.1{\scriptsize$\pm$1.3} & 40.4{\scriptsize$\pm$4.5} & 23.9{\scriptsize$\pm$10.3} & 5.1{\scriptsize$\pm$0.1} & \textbf{13.1}{\scriptsize$\pm$4.4} & 45.1{\scriptsize$\pm$0.2} & 24.9{\scriptsize$\pm$0.8} & 27.4{\scriptsize$\pm$2.6} & 33.2{\scriptsize$\pm$1.6} \\
 & \methodabbr (ours)  w/ GETA & \textbf{79.4}{\scriptsize$\pm$0.5} & \textbf{48.2}{\scriptsize$\pm$0.6} & \textbf{48.9}{\scriptsize$\pm$1.6} & \textbf{35.1}{\scriptsize$\pm$2.4} & 5.5{\scriptsize$\pm$0.3} & 10.6{\scriptsize$\pm$0.3} & \textbf{46.0}{\scriptsize$\pm$0.6} & \textbf{26.8}{\scriptsize$\pm$0.8} & \textbf{28.1}{\scriptsize$\pm$0.9} & \textbf{36.5}{\scriptsize$\pm$0.6} \\
\midrule
\multirow{3}{*}{70\%} & GETA~\cite{qu2025geta} & 41.4{\scriptsize$\pm$6.8} & 36.5{\scriptsize$\pm$1.3} & 24.9{\scriptsize$\pm$11.0} & 11.8{\scriptsize$\pm$6.0} & \textbf{4.2}{\scriptsize$\pm$0.2} & 7.1{\scriptsize$\pm$0.7} & 34.7{\scriptsize$\pm$1.1} & 19.4{\scriptsize$\pm$2.4} & 17.8{\scriptsize$\pm$5.4} & 22.0{\scriptsize$\pm$3.2} \\
 & UniQL~\cite{chiang2026uniql} & 49.2{\scriptsize$\pm$0.6} & 34.5{\scriptsize$\pm$1.8} & 18.7{\scriptsize$\pm$0.1} & 12.2{\scriptsize$\pm$3.6} & 3.8{\scriptsize$\pm$1.2} & 8.2{\scriptsize$\pm$1.1} & 35.1{\scriptsize$\pm$1.2} & 21.8{\scriptsize$\pm$3.4} & 18.4{\scriptsize$\pm$3.6} & 22.4{\scriptsize$\pm$1.5} \\
 & \methodabbr (ours)  w/ GETA & \textbf{74.5}{\scriptsize$\pm$3.1} & \textbf{38.6}{\scriptsize$\pm$1.2} & \textbf{34.0}{\scriptsize$\pm$2.0} & \textbf{36.6}{\scriptsize$\pm$0.7} & 2.7{\scriptsize$\pm$0.3} & \textbf{9.4}{\scriptsize$\pm$0.4} & \textbf{36.4}{\scriptsize$\pm$0.7} & \textbf{25.5}{\scriptsize$\pm$0.5} & \textbf{24.7}{\scriptsize$\pm$0.4} & \textbf{31.4}{\scriptsize$\pm$0.6} \\
\bottomrule
\end{tabular}}
\label{tab:plusquant_perdataset}
\end{table*}

\begin{table*}[t]
\centering
\scriptsize
\setlength{\tabcolsep}{2pt}
\caption{Per-dataset results on joint \taskabbr and token pruning from \cref{tab:plustok}. See \cref{sec:perdataset} for discussion.}
\resizebox{\textwidth}{!}{%
\begin{tabular}{llccc|cccccc|c}
\toprule
 \flops & & \multicolumn{3}{c|}{Seen} & \multicolumn{6}{c|}{Unseen} & \\
Budget & Method & CLEVRER & GQA & NExT-QA & Perc.Test & TextVQA & VizWiz & VQAv2 & VideoMME & LVB & All \\
\midrule
\multirow{3}{*}{50\%} & CoMP~\cite{wu2026comp} & 6.6{\scriptsize$\pm$2.6} & 18.9{\scriptsize$\pm$1.6} & 12.3{\scriptsize$\pm$2.7} & 20.8{\scriptsize$\pm$7.5} & 6.9{\scriptsize$\pm$3.0} & 33.7{\scriptsize$\pm$14.3} & 24.6{\scriptsize$\pm$1.4} & 17.6{\scriptsize$\pm$2.7} & 20.8{\scriptsize$\pm$2.0} & 18.0{\scriptsize$\pm$3.5} \\
 & CoMP w/ NLL & 8.7{\scriptsize$\pm$4.0} & 20.8{\scriptsize$\pm$2.8} & 20.6{\scriptsize$\pm$7.0} & 32.3{\scriptsize$\pm$7.9} & \textbf{11.6}{\scriptsize$\pm$3.4} & \textbf{36.6}{\scriptsize$\pm$5.2} & 27.6{\scriptsize$\pm$5.2} & 23.7{\scriptsize$\pm$2.6} & 26.2{\scriptsize$\pm$3.3} & 23.1{\scriptsize$\pm$2.1} \\
 & \methodabbr (ours)  w/ CoMP & \textbf{30.6}{\scriptsize$\pm$0.8} & \textbf{23.9}{\scriptsize$\pm$0.0} & \textbf{26.6}{\scriptsize$\pm$0.7} & \textbf{38.5}{\scriptsize$\pm$1.2} & 8.0{\scriptsize$\pm$0.1} & 9.7{\scriptsize$\pm$2.4} & \textbf{32.3}{\scriptsize$\pm$0.3} & \textbf{27.6}{\scriptsize$\pm$0.5} & \textbf{28.4}{\scriptsize$\pm$0.6} & \textbf{25.1}{\scriptsize$\pm$0.4} \\
\midrule
\multirow{3}{*}{70\%} & CoMP~\cite{wu2026comp} & 3.7{\scriptsize$\pm$2.3} & 0.2{\scriptsize$\pm$0.2} & 17.3{\scriptsize$\pm$3.7} & 19.4{\scriptsize$\pm$12.5} & 0.2{\scriptsize$\pm$0.2} & 0.0{\scriptsize$\pm$0.0} & 0.2{\scriptsize$\pm$0.1} & 17.1{\scriptsize$\pm$6.0} & 18.1{\scriptsize$\pm$1.7} & 8.5{\scriptsize$\pm$2.6} \\
 & CoMP w/ NLL & 6.3{\scriptsize$\pm$5.1} & 5.9{\scriptsize$\pm$0.9} & 17.0{\scriptsize$\pm$3.7} & 21.6{\scriptsize$\pm$10.2} & 1.2{\scriptsize$\pm$0.7} & 0.6{\scriptsize$\pm$0.6} & 7.6{\scriptsize$\pm$2.1} & 18.1{\scriptsize$\pm$6.6} & 18.4{\scriptsize$\pm$2.7} & 10.8{\scriptsize$\pm$2.8} \\
 & \methodabbr (ours)  w/ CoMP & \textbf{20.6}{\scriptsize$\pm$0.9} & \textbf{17.7}{\scriptsize$\pm$0.5} & \textbf{20.0}{\scriptsize$\pm$0.5} & \textbf{35.9}{\scriptsize$\pm$0.4} & \textbf{3.9}{\scriptsize$\pm$0.5} & \textbf{4.0}{\scriptsize$\pm$0.2} & \textbf{27.6}{\scriptsize$\pm$0.4} & \textbf{25.8}{\scriptsize$\pm$0.6} & \textbf{20.5}{\scriptsize$\pm$0.7} & \textbf{19.6}{\scriptsize$\pm$0.1} \\
\bottomrule
\end{tabular}}
\label{tab:plustok_perdataset}
\end{table*}

\section{Baseline Methods}
\label{sec:baseline}

\noindent\textbf{Common protocol.} Unless noted, every baseline is calibrated on the same 256 CLEVRER~\cite{yi2020clevrer} samples as \methodabbr and is applied to every block linear of both the vision and language towers. Each method's own compression knob is solved so that the whole model meets the same \flops budget $\Omega_f$ (\cref{sec:setr}). In \cref{tab:criteria}, structured pruning methods share one placement: each language-model block removes $2\,\mathrm{round}(6\rho)$ query heads (one per KV group, so GQA stays valid), each vision block removes $\mathrm{round}(16\rho)$ of its 16 heads, and every MLP loses the same fraction of channels. Only the criterion that chooses \emph{which} heads and channels to remove differs between methods. In \cref{tab:strategy} and \cref{tab:recovery}(a), methods use their own allocation.  For example, a LLM pruning method will keep the ViT dense. For unstructured pruning methods~\cite{sun2024wanda,MinSeok2026PORTA,xiong2026d2prune,zhu2018prune}, we apply their criteria to whole channels and heads rather than per-weight unless otherwise noted.
\smallskip

\noindent\textbf{FLAP~\cite{an2024flap}.} Scores each MLP channel and attention head by its fluctuation, $\mathrm{Var}(x_c)\,\|W_{:,c}\|_2^2$ on the output projection (summed over a head's columns), then creates the output-projection bias via mean contribution of removed inputs.
\smallskip

\noindent\textbf{Wanda~\cite{sun2024wanda}.} Scores weights by $|W_{ij}|\,\|X_j\|_2$, the product of its magnitude and the norm of its input, and prunes without any weight update. 
\smallskip

\noindent\textbf{Putri~\cite{coello2026putri}.} Removes MLP channels by its squared activation norm and refits the output projection sequentially through the network by least squares. Attention heads are ranked by perplexity increase after being silenced. \emph{Implementation details:} in \cref{tab:criteria} the head perplexity is computed once per head (16 calibration sequences) and ranked within each layer, replacing the paper's iterative global greedy search (unaffordable at 720 candidates), and the head unit is a single query head. In \cref{tab:strategy} we run the paper's global greedy removal of grouped heads (two per iteration) and its head-count rule. The 50\% budget is infeasible due to requiring more grouped heads than one per layer allows.
\smallskip

\noindent\textbf{Wanda~\cite{sun2024wanda}+Putri~\cite{coello2026putri}.} A hybrid we construct that uses Wanda select channels and heads and Putri's least-squares refit to update MLP layers from the unpruned model's calibration Gram (ridge $10^{-2}$ of the mean diagonal). 
\smallskip

\noindent\textbf{AMP~\cite{mugnaini2025amp}.} Scores heads by the $\ell_1$ norm of their projected output, $\|h_nW_n\|_1$, and MLP neurons by activation magnitude, then removes the same fraction of each per layer, followed by LoRA (rank 8, learning rate $3\times10^{-4}$, 2 epochs). \emph{Implementation details:} under GQA attention~\cite{ainslie2023gqa}, heads are removed in KV-group pairs and the channel fraction absorbs the rounding to meet the \flops budget.
\smallskip

\noindent\textbf{PruneLLM~\cite{yang2025prunellm}.} Scores channels by information entropy, selects them with K-means clustering, and allocates sparsity across blocks, sub-layers, KV groups and channels by transfer entropy. Bias compensation uses the signed calibration mean, and no fine-tuning follows. 
\smallskip

\noindent\textbf{D$^2$Prune~\cite{xiong2026d2prune}.} Scores weights with a dual-Taylor expansion over both input and output activations plus an OBS term from SparseGPT~\cite{frantar2023sparsegpt} to estimate reconstruction error. \emph{Implementation details:} we use the paper's default hyperparameters ($\lambda_1{=}1$, $\lambda_2{=}0$, $s{=}1500$) to apply the OBS update sequentially by layer. 
\smallskip

\noindent\textbf{PORTA~\cite{MinSeok2026PORTA}.} Scores each weight by its input-feature activation \emph{variance} times its magnitude, $v_j|W_{ij}|$, and sets per-layer sparsity from a variance-weighted layer score. 
\smallskip

\noindent\textbf{SVD-LLM~\cite{wang2025svdllm}.} This \taskabbr method uses truncation-aware whitening where each matrix selects factors in the Cholesky-whitened space, $W\approx(WL)_rL^{-1}$ (more details in \cref{sec:balanced}). \emph{Implementation details:} in \cref{tab:criteria}, \cref{tab:architectures} and \cref{tab:recovery}(b) every block linear of both towers is factorized, and the closed-form layer-wise update of the preprint is applied at ratios of 40\% and above (our 50\% budget only). In \cref{tab:strategy} and \cref{tab:recovery}(a) we follow their published work: language model only, no update without recovery (its SVD-LLM (W) ablation), and its sequential low-rank update (LoRA on $U$ with $V$ frozen, then on $V$) as recovery. 
\smallskip

\noindent\textbf{Basis Sharing~\cite{wang2025basissharing}.} Shares one whitened low-rank basis across a pair of consecutive layers, with per-layer coefficients, plus SVD-LLM's update at ratios of 40\% and above. \emph{Implementation details:} only gate/up are shared across layer pairs because sharing q/k/v increased activation error 2--3$\times$. Layers 0--1 are not shared because their Gram traces differ by 21.6$\times$. After the joint fit, each member's coefficients are re-solved against its own Gram, and the update uses a $10^{-4}$ relative ridge instead of a plain inverse. The vision tower is factorized but not shared. 
\smallskip

\noindent\textbf{LLM-Pruner~\cite{ma2023llmpruner}.} Removes coupled structure groups (heads, MLP channels) ranked by first-order plus Fisher Taylor importance, summed over each group.  \emph{Implementation details:} it keeps the first three blocks and the last block dense, applies a uniform ratio to the rest, and recovers with LoRA (rank 8, $\alpha{=}16$, dropout 0.05, learning rate $10^{-4}$, 2 epochs). Heads are rounded to KV-group pairs and the channel count is solved to meet the \flops budget. 
\smallskip

\noindent\textbf{SlimLLM~\cite{guo2025slimllm}.} Selects heads by a Pearson-correlation score with a greedy swap, and channels by a score in the principal-component basis of the centred output. It sets per-layer ratios with its Eq.~11, skipping the first and last layers, refits o\_proj and down\_proj by per-output linear regression, and recovers with LLM-Pruner's LoRA. \emph{Implementation details:} the mean ratio is bisected to meet the \flops budget. $\alpha$ is interpolated between the paper's two stated points (10 at 20\%, 7 at 50\%). Head selection respects GQA attention groups.
\smallskip

\noindent\textbf{AdaSVD~\cite{Li_2026_CVPR_Adasvd}.} Extends SVD-LLM with adaComp, which alternately re-solves $U$ and $V$ by pseudo-inverse against the deviated inputs (one iteration), and adaCR, which sets per-block retention from block importance (span 0.1). \emph{Implementation details:} we use the exact Gram over all calibration samples rather than the paper's stack-of-batch approximation. Block importance is $1-\cos(\text{block input},\text{block output})$ as in their paper's Fig.~3 and is normalized per tower. The retention ratio is bisected to the \flops budget, and the ViT is also compressed.
\smallskip

\noindent\textbf{Cat~\cite{lu2026skipcat}.} Uses the same shared-input \taskabbr for twin layers as our work (both our work and theirs will still factorize singletons). \textbf{Cat w/Skip} additionally writes every input factor in block-skip form $V_1[I\;|\;V_1^{-1}V_2]P^\top$, solving rank against the $r(n{+}m{-}r)$ cost. Columns are chosen by maxvol (entries of $|V_1^{-1}V_2|\le1.05$). \emph{Combining with \methodabbr:} Skip changes only how each factor is stored, so it composes directly with our factorization. We fit the factors with \methodabbr, then rewrite every input factor, shared or not, in the same block-skip form with the same maxvol column selection, and solve the per-layer rank against the block-skip cost so the \flops budget is met exactly.
\smallskip

\noindent\textbf{Minitron~\cite{muralidharan2024minitron}.} Ranks heads and MLP neurons by activation importance alone: the mean over samples of each sample's token-wise $\ell_2$ norm. Recovery is pure logit distillation from the dense model ($\alpha{=}1$, $T{=}1$, full fine-tune). 
\smallskip

\noindent\textbf{RLP~\cite{lu2026reassessing}.} A recovery method that drops the last blocks and then fine-tunes only the lm\_head and the last three remaining language-model blocks.
\smallskip

\noindent\textbf{INTERLACE~\cite{madinei2026interlace}.} Walks the network in non-overlapping triplets of consecutive layers. In each, it drops the more locally redundant of the first two ($1-\cos$ of block input and output), fine-tunes the survivor, and freezes the third as an anchor. The 50\% \flops budget is infeasible since removing one layer per triplet caps the reduction.
\smallskip

\noindent\textbf{Structured GMP~\cite{zhu2018prune}.} Gradual magnitude pruning with a cubic sparsity schedule, starting from the dense model during fine-tuning. \emph{Implementation details:} it uses 20 pruning steps over the first two-thirds of training. 
\smallskip

\noindent\textbf{RCPU~\cite{haruta2026rcpu}.} Scores each input column $j$ of an output projection by $\|W_{:,j}\|_2\,\|X_j\|_2\,\mathrm{Var}(X_j)$, \ie, Wanda-sp's score weighted by the activation variance, and removes the lowest-scoring MLP channels and heads (a head's score is the sum over its columns). It then compensates the pruning error with a rotation rather than a least-squares refit: the kept columns are updated as $Q^\ast W_S$, where $Q^\ast=UV^\top$ is the orthogonal Procrustes solution aligning the pruned output $Z=W_SX_S$ with the original output $Y=WX$, from the SVD $YZ^\top=U\Sigma V^\top$. Both corrections are calibration-only and applied sequentially through the already-pruned prefix. Attention is pruned first followed by the MLP on the corrected attention's output.
\smallskip

\noindent\textbf{TRSP~\cite{feng2026trsp}.} Performs depth pruning with two regularization stages on the calibration set. First, a learnable scalar $s_i$ scales each language-model layer, and the model is trained on its language-modeling loss plus $\lambda_1\sum_i|s_i|$. After each pass the layer with the smallest $|s_i|$ is masked until the target number of layers is selected. Second, with the selected layers active again, training adds $\lambda_2\sum_{i\in P}\|X^i_{\mathrm{out}}-X^i_{\mathrm{in}}\|$, pushing those layers toward the identity so their knowledge moves to the kept layers before they are removed. \emph{Implementation details:} we use the paper's $\lambda_1{=}5{\times}10^{-3}$, $\lambda_2{=}10^{-3}$ and optimizer (AdamW, learning rate $2{\times}10^{-5}$, all language-model weights trainable). Two changes were needed for the scalars to select layers in a pre-norm model. Specifically, when applied to the full layer output, as in the paper, the scalars barely moved from their initialization at the paper's learning rate. Additionally, they selected the first and last layers, because scaling the residual stream is largely undone by the following normalization. Thus, we apply $s_i$ to the residual branch, $X^i_{\mathrm{out}}=X^i_{\mathrm{in}}+s_i f_i(X^i_{\mathrm{in}})$, so $s_i{=}0$ is exactly the removed layer, and train the scalars with a learning rate of $10^{-2}$ (the weights keep $2{\times}10^{-5}$). For InternVL3-2B~\cite{zhu2025internvl3}, we remove the closest whole-layer match for the 10\%/30\%/50\% \flops budgets, resulting in 3/10/16 of the 28 layers removed.
\smallskip

\noindent\textbf{DDP~\cite{huang2026ddp}.} In this work the pretrained weights are frozen and only a real-valued gate $m_k=\mathrm{ReLU}(z_k)$ on every attention head and MLP channel is trained. A deterministic retention surrogate that is sharpened over training ($\mu_t$ annealed from 0.5 to 0.05) is used by an augmented-Lagrangian sparsity penalty and a binarization term. The loss is CE $+\,2\,\mathrm{KL}$ to the dense model, plus these terms. \emph{Implementation details:} training uses the paper's settings (AdamW, learning rate $2{\times}10^{-2}$ for the gates, global batch 16, warm-up then cosine decay). The global budget is enforced separately for each of the four module types (language-model and vision heads and channels), each at the shared placement's keep ratio, so the \flops match the other rows exactly. The final kept set is the top gates of each type, with at least one unit per layer and an equal number of query heads per KV group under GQA. The remaining units are removed and the kept gates are folded into the output projections. Gates are trained on our SFT data with cached teacher logits. In our experiments some run's learned gates removed two first-layer query heads that the model cannot function without (removing the same two heads from a healthy seed drops accuracy to near zero, and restoring either recovers most of it). Thus, we exclude the first language-model layer's heads from gating and pruning and remove its planned heads from the other layers instead, which keeps the head count and \flops unchanged.

\noindent\textbf{CoMP~\cite{wu2026comp}.} This method jointly removes MLP channels in both towers and visual tokens inside the language model, chosen greedily stage by stage. Channels are scored by a token-weighted Wanda criterion, $\|W_{:,i}\|_1\,(\sum_n \omega_n x_{n,i}^2)^{1/2}$, where $\omega_n$ is the attention token $n$ receives (maximum over heads, summed over queries), so tokens already pruned stop contributing.  \emph{Implementation details:} Tokens are dropped in a cascade after language-model blocks 8, 14 and 20 by the same attention score. Each stage applies one mode (5\% of a tower's MLP channels, or 15\% of the kept tokens), and the next mode is the one with the lowest smoothed cost $\Delta\text{accuracy}/\Delta\text{\flops}$, with interval-weighted random exploration (the paper's $\rho{=}0.2$, $\tau{=}5$, $\lambda_0{=}0.4$, $I_{\max}{=}5$). The search runs until the \flops budget is reached, shrinking the final stage to land on it exactly, with at least 1/8 of the image tokens kept. \emph{Combining with \methodabbr:} we keep CoMP's token pruning and replace its weight pruning with our factorization. The token rule CoMP reaches (keep ratio, drop blocks and attention criterion) is applied to the dense model. Its \flops savings are charged first, and \methodabbr factorizes every linear layer of both towers to cover the rest of the budget. The activation statistics for the factorization are collected with the token cascade active, leaving out of each language-model layer's statistics the image tokens already dropped before that block. Thus, the factors are fitted to the activations the token-pruned model actually processes.
\smallskip

\begin{table*}[t]
\centering
\caption{Number of sample questions per split. The recovery set is the union of the three training splits
(134{,}021 question--answer pairs).}
\begin{tabular}{lllrrl}
\toprule
& Dataset & Media & Train & Val & Test (source) \\
\midrule
\multirow{3}{*}{Seen datasets} & CLEVRER~\cite{yi2020clevrer}   & video & 61{,}064 & 14{,}235 & 10{,}000 (subset of official val) \\
& GQA~\cite{hudson2019gqa}       & image & 65{,}093 & 25{,}803 & 12{,}578 (testdev-balanced) \\
& NExT-QA~\cite{xiao2021nextqa}   & video &  7{,}864 &     896  &  4{,}221 (video-level re-split) \\
\midrule
\multirow{6}{*}{Unseen datasets} & Perception Test~\cite{patraucean2023perception} & video & -- & -- & 19{,}140 (val) \\
& VQAv2~\cite{goyal2017vqav2}     & image & -- & -- & 10{,}000 (subset of val) \\
& VizWiz~\cite{gurari2018vizwiz}    & image & -- & -- &  4{,}319 (val) \\
& TextVQA~\cite{singh2019textvqa}   & image & -- & -- &  5{,}000 (val) \\
& Video-MME~\cite{fu2025videomme} & video & -- & -- &  2{,}700 (w/o subtitles) \\
& LongVideoBench~\cite{wu2024longvideobench} & video & -- & -- & 1{,}337 (val) \\
\bottomrule
\end{tabular}
\label{tab:datasets}
\end{table*}

\noindent\textbf{CoMP w/NLL.} Our extension of CoMP's search with two changes. First, it adds a ViT attention-head mode (one head per stage, scored by the same token-weighted criterion on the head's output-projection columns, at least two heads kept). We do not add language-model heads since removing even one query head per KV group from every block collapses accuracy under every head score we tried. Second, it scores each stage by the negative mean log-likelihood of the dense model's own 16-token answers on the calibration samples, teacher-forced, instead of accuracy. Calibration accuracy falls close to zero during the search, after which accuracy differences between stages are mostly noise. 
\smallskip

\noindent\textbf{GETA~\cite{qu2025geta}.} This method performs structured pruning and quantization-aware training. Each layer gets a learnable weight quantizer (step $d$, clip $q_m$ and exponent $t$, which together set its bit-width), and the QASSO optimizer trains in four stages. After warm-up, a projection stage lowers the allowed bit-width each period and projects $d$ into that range. Then the joint stage picks redundant channel and head groups by HESSO saliency each period and decays them to zero while their quantizers keep training. A cool-down step fixes each layer's rounded bit-width. \emph{Implementation details:} we follow the authors' released code. The pruning groups are LLM MLP channels, one query head per KV group, ViT heads and ViT MLP channels, and groups are removed until the \flops budget is met rather than to a fixed group count. Every linear layer of both towers is quantized and ends at exactly 4 bits (embeddings, language-model head and projector stay in bf16). \emph{Combining with \methodabbr:} We start from the calibration-only \methodabbr model and run QASSO with pruning disabled. Its warm-up and projection stages bring every factor to 4 bits, and the cool-down holds the rounded bit-widths for the rest of training. Each factor gets its own quantizer, and a shared input factor is quantized once for all the layers that use it. 
\smallskip

\noindent\textbf{UniQL~\cite{chiang2026uniql}.} Performs structured pruning and 4-bit quantization that serves several budgets from one model. First, channels and head dimensions are reordered by importance with exact, output-preserving transforms.  For example, MLP channels use ridge leverage of their activations, query/key dimensions by their activation energy, and value/output projections by an activation-whitened SVD. Each layer's pruning rate is set from its block influence, $1-\cos(\text{input},\text{output})$, through a softmax. A masked LoRA fine-tune samples one budget per step and masks what that budget removes, then the model is quantized with GPTQ (4 bits, groups of 128) and truncated to each budget. \emph{Implementation details:} we follow the paper's recipe (LoRA rank 8 on every linear layer of both towers, 5 epochs, learning rate $10^{-4}$). The shared rate scale is bisected so each budget meets its \flops target, and kept sizes are multiples of 8. Quantization covers the same layers as GETA.

\section{Dataset Statistics}
\label{sec:datasets}

\begin{table*}[t]
\centering
\small
\setlength{\tabcolsep}{3pt}
\caption{Per-dataset results for \cref{tab:base}.}
\begin{tabular}{lccc|cccccc|c}
\toprule
 & \multicolumn{3}{c|}{Seen} & \multicolumn{6}{c|}{Unseen} & \\
 Method & CLEVRER & GQA & NExT-QA & Perc.Test & TextVQA & VizWiz & VQAv2 & VideoMME & LVB & All \\
\midrule
InternVL3-1B~\cite{zhu2025internvl3} & 36.1 & 57.2 & 70.9 & 51.7 & 72.5 & 49.3 & 75.1 & 40.7 & 41.3 & 55.0 \\
InternVL3-2B~\cite{zhu2025internvl3} & 36.5 & 59.3 & 75.3 & 55.1 & 77.3 & 53.2 & 78.2 & 46.2 & 46.8 & 58.7 \\
Qwen2-VL-2B~\cite{wang2024qwen2vl} & 38.8 & 60.0 & 83.9 & 47.7 & 70.2 & 64.3 & 78.7 & 45.0 & 43.4 & 59.1 \\
Qwen3-VL-2B~\cite{bai2025qwen3vl} & 35.6 & 58.6 & 69.0 & 51.0 & 70.6 & 70.6 & 77.1 & 46.3 & 44.2 & 58.1 \\
Qwen3.5-2B~\cite{qwen2026qwen35} & 39.2 & 56.9 & 63.4 & 50.2 & 69.0 & 64.0 & 77.9 & 47.8 & 45.7 & 57.1 \\
Gemma4 E2B~\cite{gemmateam2026gemma4} & 23.1 & 41.8 & 47.8 & 47.4 & 65.3 & 41.9 & 56.7 & 40.9 & 36.2 & 44.6 \\
\midrule
InternVL3-2B + SFT & 81.8{\scriptsize$\pm$0.8} & 63.1{\scriptsize$\pm$0.1} & 78.1{\scriptsize$\pm$0.2} & 56.6{\scriptsize$\pm$0.4} & 73.5{\scriptsize$\pm$0.5} & 53.3{\scriptsize$\pm$0.7} & 73.8{\scriptsize$\pm$0.1} & 43.8{\scriptsize$\pm$0.3} & 46.8{\scriptsize$\pm$0.6} & 63.4{\scriptsize$\pm$0.1} \\
\bottomrule
\end{tabular}
\label{tab:base_perdataset}
\end{table*}

\begin{table}[t]
    \centering
    \caption{Base architecture performance to provide an oracle reference for each setting in our experiments. We report results for (top) off-the-shelf and (bottom) with Supervised FineTuning (SFT) over three random seeds.}
    \begin{tabular}{lccc}
    \toprule
    Model & Seen & Unseen & All\\
    \midrule
    InternVL3-1B~\cite{zhu2025internvl3} & 54.7 & 55.1 & 55.0\\
    InternVL3-2B~\cite{zhu2025internvl3} & 57.0 & 59.5 & 58.7\\ 
    Qwen2-VL-2B~\cite{wang2024qwen2vl} & 60.9 & 58.2 & 59.1\\
    Qwen3-VL-2B~\cite{bai2025qwen3vl} & 54.4 & 60.0 & 58.1\\
    Qwen3.5-2B~\cite{qwen2026qwen35} & 53.1 & 59.1 & 57.1\\
    Gemma4 E2B~\cite{gemmateam2026gemma4} & 37.5 & 48.1 & 44.6\\
    \midrule
    InternVL3-2B + SFT & 74.3{\scriptsize$\pm$0.2} & 58.0{\scriptsize$\pm$0.2} & 63.4{\scriptsize$\pm$0.1} \\
    \bottomrule
    \end{tabular}
    \label{tab:base}
\end{table}

In this paper we focus our evaluation on settings built for speed where there are few other alternatives.  Thus, many of the backbones we use are the smallest (or close to the smallest) in their model families (more discussion in \ref{sec:base_arch}), but it also means we process videos for low latencies to replicate a setting where models are deployed at scale.
Specifically, following~\cite{kim2024igvlm,chen2024sharegpt4video}, videos are presented as a single keyframe grid of 16 frames sampled uniformly over the clip, tiled $4\times4$ at $224$\,px with 4\,px white separator bands, and resized to one $448\times448$. Below we provide a high-level description of each dataset, see \cref{tab:datasets} for a summary.
\smallskip

\noindent\textbf{Seen datasets.}
\textit{CLEVRER}~\cite{yi2020clevrer} contains synthetic collision videos with descriptive,
explanatory, predictive and counterfactual questions. Descriptive questions are open-ended; the
other three types are multiple-select. The official test split ships no answers, so we use the
official validation split as our test set and carve our validation set out of official train by
video (10\% of videos).
\textit{GQA}~\cite{hudson2019gqa} is compositional question answering over real images. We draw
train and validation from the official balanced train and validation splits (subsampled by
image) and use the balanced \textsc{testdev} split as test.
\textit{NExT-QA}~\cite{xiao2021nextqa} is multiple-choice causal and temporal reasoning over
everyday videos. We pool the annotated videos and split them by video into train, validation and
test (900 / 100 / 490 videos), so the same video never appears in two splits.  The numbers reported in \cref{tab:datasets} reflect the number of questions (\ie, videos can have multiple questions).
\smallskip

\noindent\textbf{Unseen datasets.}
\textit{Perception Test}~\cite{patraucean2023perception} (multiple-choice video, \texttt{valid}
release, since test answers are not public), \textit{VQAv2}~\cite{goyal2017vqav2},
\textit{VizWiz}~\cite{gurari2018vizwiz} and \textit{TextVQA}~\cite{singh2019textvqa} (open-ended
image question answering, validation sets, ten annotator answers each),
\textit{Video-MME}~\cite{fu2025videomme} (multiple-choice, without subtitles) and
\textit{LongVideoBench}~\cite{wu2024longvideobench} (multiple-choice, validation split, the only
split with public answers).
\smallskip

\noindent\textbf{Splits construction.}
All splits are built deterministically from a fixed hash salt, and every split is disjoint from
the others at the level of the underlying video or image, not only the question. Calibration data
for data-aware compression is drawn from 256  CLEVRER training samples. Validation splits are used for selecting hyperparameters for each method. Test splits are used only for reporting.

\section{Base Architecture Discussion}
\label{sec:base_arch}

Each model we use are (one of) the smallest in its architecture family.  These are chosen deliberately as they reflect likely applications of this type of computational reduction work.  For example, if we were to compress a 72B model by 50\%, one could simply argue that you could use a 32B model from the same family for similar benefits instead.  Further, our experiments with InternVL3-1B and InternVL3-2B show that two models of different sizes from the same family can behave very differently.  For example, \cref{fig:gamma} shows the 1B variant was more sensitive to choice of $\gamma$ than the 2B model.  Thus, evaluating at different scales would not necessarily replicate our setting where few alternatives exist other than to use \taskabbr or methods like it. Further, compact models also may be deployed to edge devices, where reducing memory and latency is even more important. This leads us to focus on these smaller architectures, with the InternVL3 evaluation providing a snapshot of model family variations.
\smallskip

\noindent\textbf{Dense model results.} \cref{tab:base,tab:base_perdataset} provide the overall and per-dataset results, respectively, for the dense (unpruned/factorized) models as a reference point.  In some cases the Unseen datasets get higher zero-shot performance than those of the Seen datasets.  Thus, higher Unseen results by pruning/\taskabbr methods in the other results tables does not necessarily speak to better generalization since its peak performance is simply higher for some checkpoints.